\documentclass[lettersize,journal]{IEEEtran}
\usepackage{amsmath,amsfonts}
\usepackage{algorithmic}
\usepackage{algorithm}
\usepackage{array}
\usepackage[caption=false,font=normalsize,labelfont=sf,textfont=sf]{subfig}
\usepackage{textcomp}
\usepackage{stfloats}
\usepackage{url}
\usepackage{verbatim}
\usepackage{graphicx}
\usepackage{multicol}
\usepackage{multirow}
\usepackage{makecell} 
\usepackage{cite}
\usepackage{hyperref}
\usepackage{xcolor}

\usepackage{color,soul}
\sethlcolor{yellow} 
\usepackage{empheq}

\usepackage[many]{tcolorbox}
\usepackage{ctable}

\tcbset{
  myhlight/.style={
    colback=yellow,
    arc=0pt,
    outer arc=0pt,
    boxrule=0pt,
    top=2pt,
    bottom=2pt,
    left=2pt,
    right=2pt,
  },
  highlight math style={myhlight}
}

\newtcolorbox{myhl}{
  breakable,
  myhlight
}

\begin{document}

\title{Advancing Adversarial Training by Injecting Booster Signal}

\author{Hong Joo Lee, Youngjoon Yu, and Yong Man Ro,~\IEEEmembership{Senior Member,~IEEE,}
\thanks{This work was supported by Center for Applied Research in Artificial Intelligence (CARAI) grant funded by DAPA and ADD (UD190031RD).}
\thanks{H. J. Lee, Y. Yu, and Y. M. Ro are with the Image and Video Systems
Laboratory, School of Electrical Engineering, Korea Advanced Institute of
Science and Technology (KAIST), Daejeon, 34141, South Korea (e-mail:
dlghdwn008@kaist.ac.kr; greatday@kaist.ac.kr; ymro@kaist.ac.kr). \textit{(Corresponding author: Yong Man Ro.)}}}



\maketitle

\begin{abstract}
Recent works have demonstrated that deep neural networks (DNNs) are highly vulnerable to adversarial attacks. To defend against adversarial attacks, many defense strategies have been proposed, among which adversarial training has been demonstrated to be the most effective strategy. However, it has been known that adversarial training sometimes hurts natural accuracy. Then, many works focus on optimizing model parameters to handle the problem. Different from the previous approaches, in this paper, we propose a new approach to improve the adversarial robustness by using an external signal rather than model parameters. In the proposed method, a well-optimized universal external signal called a booster signal is injected into the outside of the image which does not overlap with the original content. Then, it boosts both adversarial robustness and natural accuracy. The booster signal is optimized in parallel to model parameters step by step collaboratively. Experimental results show that the booster signal can improve both the natural and robust accuracies over the recent state-of-the-art adversarial training methods. Also, optimizing the booster signal is general and flexible enough to be adopted on any existing adversarial training methods.
\end{abstract}


\section{Introduction}
\IEEEPARstart{D}{espite} the phenomenal success of deep neural networks (DNNs) in various applications such as computer vision \cite{he2016deep,lee2020structure,lin2017feature,kim2019bbc}, audio recognition \cite{moritz2020streaming,kim2021cromm,kim2021lip,dong2018speech}, and natural language processing \cite{zhang2020text,ma2019tensorized,tenney2019bert}, they are highly vulnerable to adversarial examples \cite{8611298,akhtar2021advances,zhang2021adversarial,9302639}. By adding small and imperceptible perturbation to input data, it changes the original prediction \cite{goodfellow2014explaining,madry2017towards,carlini2017towards}. The adversarial examples have imposed serious threats to safety-related applications such as autonomous driving cars and medical diagnosis. Therefore, it is necessary to develop defense strategies against adversarial attacks.

To mitigate the vulnerability of DNNs, many defense methods such as input pre-processing based defenses \cite{das2017keeping,liu2019feature,naseer2020self,Mao_2021_ICCV,naseer2019local} and randomization \cite{raff2019barrage,xie2017mitigating,lee2020robust,liu2018towards}, have been proposed. However, they are easily broken in white-box attack settings \cite{athalye2018obfuscated} since their defensive capability originates from gradient masking. 

Among the various defense methods \cite{Akhtar_2018_CVPR,pmlr-v139-zhao21e,srinivasan2021robustifying,9466420}, Adversarial Training (AT) has been demonstrated to be the most effective defense strategy \cite{athalye2018obfuscated,bai2021recent}. They train DNNs with adversarial examples by solving min-max optimization problems \cite{madry2017towards,zhang2021geometryaware,zhang2020attacks,Wang2020Improving,pmlr-v97-zhang19p,kannan2018adversarial}. The adversarial example is generated by maximizing the loss value, and the parameters of the DNNs are optimized to minimize the loss value against adversarial examples. Although many adversarial training methods have improved adversarial robustness, it has a critical problem that hurts natural accuracy (test on clean example) \cite{pmlr-v97-zhang19p}. To release the problem, some works tried to improve the natural accuracy while maintaining the adversarial robustness or improve the robustness while maintaining the natural accuracy \cite{Wang2020Improving,zhang2021geometryaware}. Furthermore, recently, improving both robustness and natural accuracies attracts more interest \cite{zhang2020attacks,rade2021helper}. Most of these methods tried to optimize the model parameters with well-designed loss functions to improve adversarial robustness.

\begin{figure}[t]
	\centering
	    \includegraphics[width=0.98\linewidth]{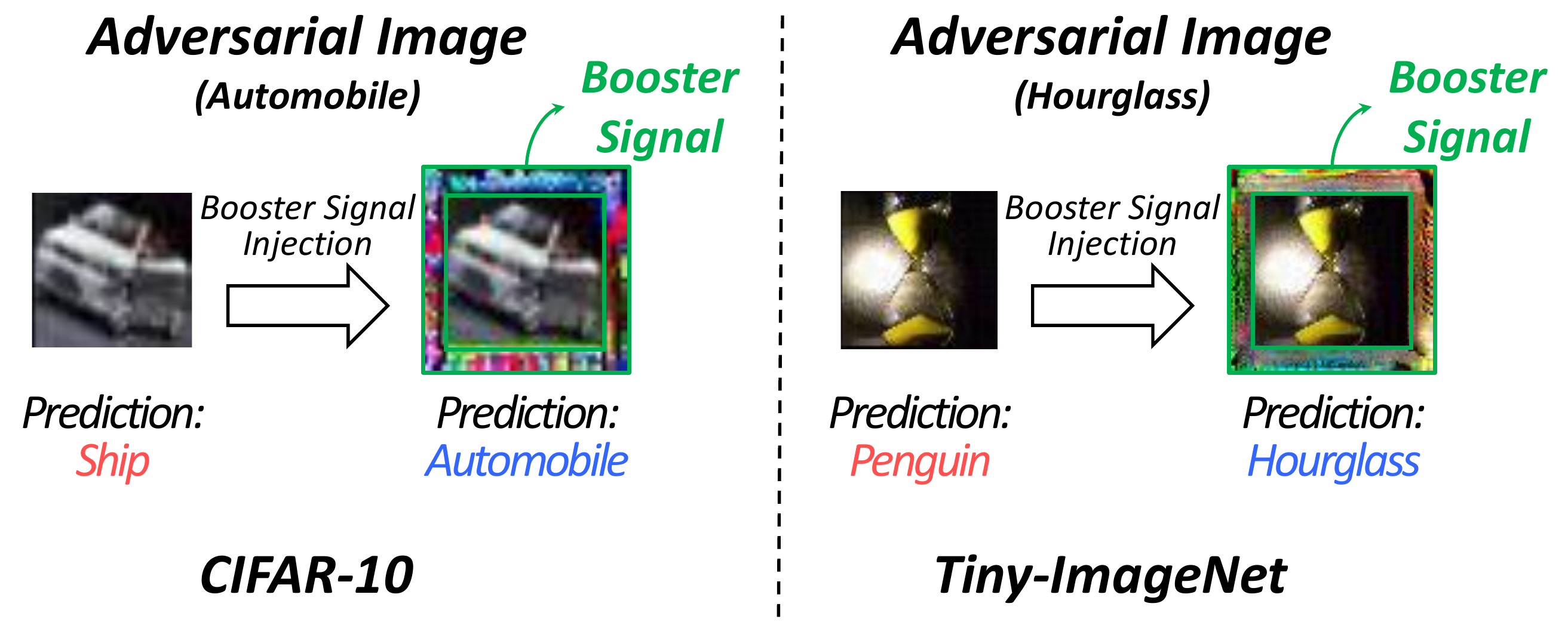}
     \vspace{-0.3cm}
	\caption{Concept of the proposed booster signal. By injecting a booster signal into the outside of the image, it can defend against the adversarial attack. Note that even though the adversarial examples are generated on the booster signal injected image, it could correctly classify the input.} 
 \vspace{-0.5cm}
\label{fig:1}
\end{figure}

Breaking away from the existing approaches that only handle model parameters, we propose a new perspective for improving adversarial robustness and natural accuracy by an external signal. The motivation of the proposed method is raised by the following questions:
\begin{center}
``\textit{Is it possible to improve the adversarial robustness and natural accuracy through the external signal other than model parameters? If possible, can we boost the adversarial robustness by collaborating the external signal and existing adversarial training methods?}"
\end{center}

To answer the aforementioned question, we investigate this intriguing, yet thus far-overlooked aspect of the external signal. We consider the external signal as a signal injected into the input data and find that injecting a well-optimized external signal reduces the gradient of the cost function with respect to the input data. Then, it makes the input be robust against the adversarial attack and improves natural accuracy. Since the booster signal is a separate signal independent of the model parameters, it could improve robustness by applying it to any existing AT methods collaboratively.


Fig. \ref{fig:1} briefly illustrates the concept of the booster signal. As shown in the figure, the booster signal is placed on the outside of the input image so that the booster signal and input image do not overlap. Then, even when the input is misclassified by adversarial perturbation, the injected booster signal serves to correctly classify the input by injecting the booster signal. 

In this paper, we propose a novel framework to optimize the external signal and DNNs collaboratively. In the proposed framework, the booster signal and DNNs are optimized over 4 steps. In the first step, we optimize the model with adversarial perturbations as previous adversarial training methods have done. This step makes DNNs have robust decision boundaries as previous adversarial training has done. Then, in the second step, we optimize the booster signal with a clean image set that represents the training data distribution well. Optimizing the booster signal for each individual image is challenging because the ground-truth label is unknown during the inference time. Therefore, we optimize the booster signal that can be applied to any input image. By optimizing the booster signal through the whole image sets, the booster signal can correctly classify almost all images in the data distribution. In the third step, we optimize the booster signal with adversarial examples in an adversarial way. When generating an adversarial example, in the third step, we use the booster signal injected image. Therefore, the booster signal is optimized to defend against those adversarial examples. Since the booster signal is optimized to defend against adversarial perturbation that attacks the booster signal injected input, it can be effective under white-box attack settings and does not suffer from the gradient masking phenomenon. Through steps 2 and 3, the booster signal reduces the gradient of the cost function with respect to the input data and makes the input itself becomes robust against adversarial attacks. Finally, in the fourth step, we conduct existing adversarial training methods with the booster signal injected images to fit the new data distribution induced by the booster signal injection. We repeat the aforementioned optimization steps for every epoch. Then, during the inference, we inject the optimized booster signal to the outside of the image and feed-forward it to the model.

To conclude the introduction, we outline the major contributions of this work as follows:

\begin{itemize}
    \item We introduce the booster signal that can improve both the natural and robust accuracies in AT methods. This is the first approach to improve adversarial robustness by optimizing an external signal in AT methods.
    \item The booster signal is image agnostic that could be effective regardless of input images. Therefore, once the booster signal is optimized, we can inject the booster signal into any input image and improve both natural accuracy and robust accuracy.
    \item Since the booster signal is separated from the model parameters, it can be applied in parallel with the existing AT method. Experimental results show that optimizing the booster signal is general and flexible enough to be adopted on any existing AT methods.
\end{itemize}

\section{Related Work}
\subsection{Adversarial Attack}
It has been widely known that DNNs are highly vulnerable to adversarial perturbations \cite{8611298,akhtar2021advances,zhang2021adversarial,9302639}. By adding small and imperceptible perturbations into input data, it misleads the DNN predictions \cite{li2020vulnerability,9308597,srinivasan2019black,duan2021advdrop,wang2020hamiltonian}. Fast Gradient Sign Method (\textbf{FGSM}) \cite{goodfellow2014explaining} is a simple and effective adversarial attack method. It generates adversarial perturbation by using the gradient of the loss function with respect to the input data at once. As an extension of FGSM, Projected Gradient Descent (\textbf{PGD}) is proposed. It iteratively updates adversarial perturbations with a small step size. It also uses the gradient of the loss function with respect to the input data. Carlini \& Wagner (\textbf{C\&W}) \cite{carlini2017towards} attack explores an optimization-based adversarial attack method. It optimizes adversarial perturbations that change the logit values with minimal distortion. Recently, a more strong attack called \textbf{AutoAttack} \cite{croce2020reliable} has been proposed. It attacks by ensembling four adversarial attacks including APGD-CE, APGD-DLR, FAB \cite{croce2020minimally} and Square attack \cite{andriushchenko2020square}.APGD-CE and APGD-DLR are automatized variants of the PGD attack proposed in \cite{croce2020reliable}. They generate adversarial perturbation by using a step-learning rate schedule adaptively. Since AutoAttack is a powerful attack, it is used as a benchmark for evaluating robustness \cite{croce2021robustbench}.

\begin{figure*}[!t]
	\centering
	    \includegraphics[width=0.95\linewidth]{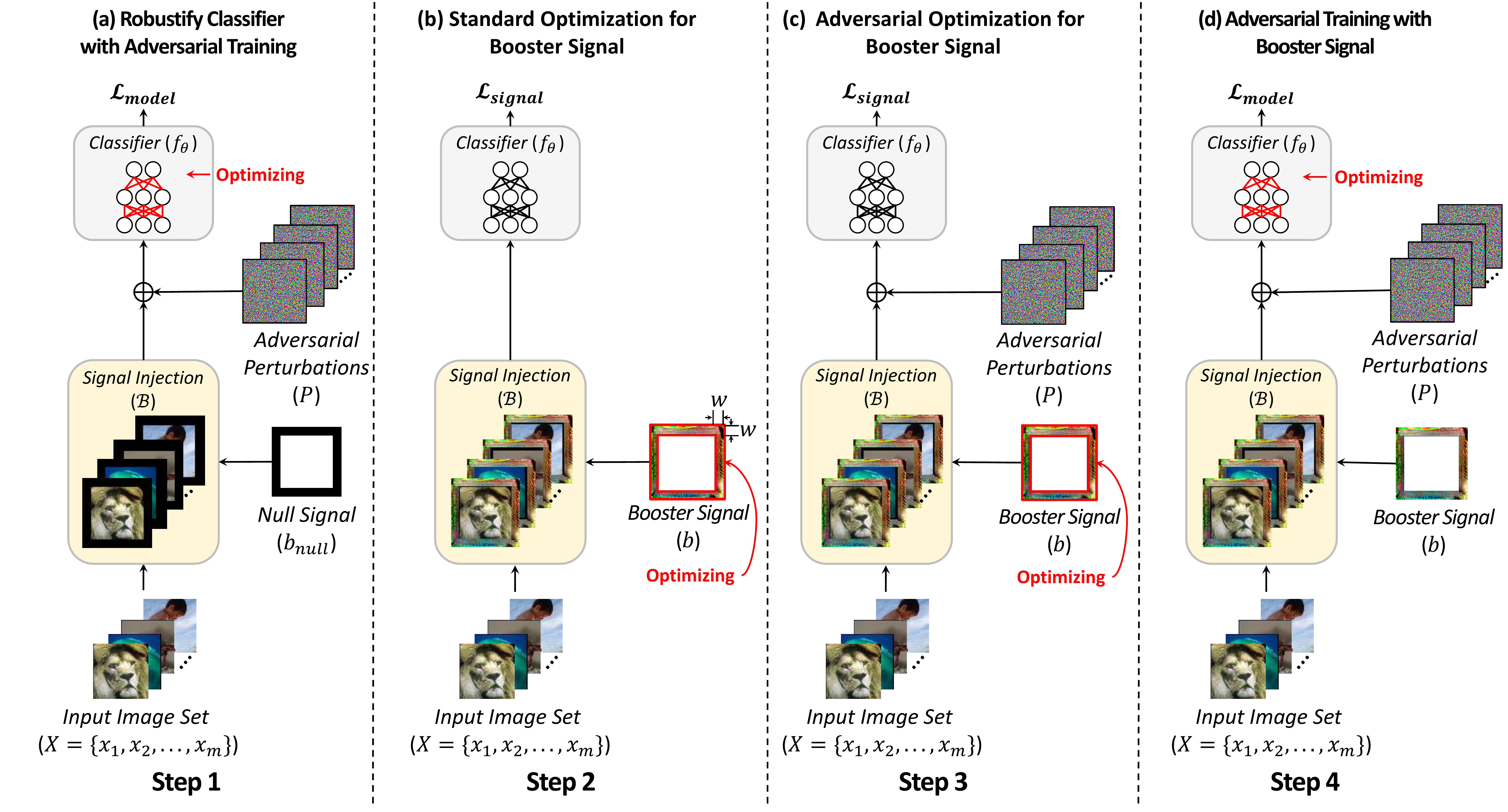}
	\caption{Overview of the proposed optimization process for the booster signal and model parameters. The classifier and booster signal are optimized step by step collaboratively. Red lines and square boxes denote the optimizing model parameters and booster signal respectively.} 

	\label{fig:2}
\end{figure*}

\subsection{Defense: Adversarial Training}
Adversarial Training (AT) is known as the most effective approach to defend against adversarial attack \cite{athalye2018obfuscated,bai2021recent}. By solving a min-max optimization problem between model parameters and adversarial perturbation, it improves the adversarial robustness. It can be formulated as follows: 
\begin{equation}
\begin{aligned}
            & \underset{\theta}{\textrm{argmin}}\mathcal{L}_{model}(f_{\theta}(x+p^{adv}),y), \\ 
            \textrm{where}& \quad    p^{adv}=\underset{||p||<\epsilon}{\textrm{argmax}}\mathcal{L}_{adv}(f_{\theta}(x+p),y),
\end{aligned}
\label{eq:1}
\end{equation}
where $x$ is the input, $p$ and $p^{adv}$ are the adversarial perturbations, $y$ is the ground truth class of input $x$, $f_{\theta}$ is the output of a model with parameter $\theta$, $\mathcal{L}_{adv}$ is the loss for generating adversarial perturbation, $\epsilon$ denotes the perturbation budget and $\mathcal{L}_{model}$ is the loss for optimizing the parameters of the model. Following Eq. \ref{eq:1}, many variants of adversarial training methods have been proposed by designing $\mathcal{L}_{model}$ and $\mathcal{L}_{adv}$.

\noindent \textbf{Madry \cite{madry2017towards}:} Madry et al. proposed a multi-step gradient-based attack known as PGD attack method and improved adversarial robustness by training the model with PGD perturbations. They used $\mathcal{L}_{model}$ and $\mathcal{L}_{adv}$ as Cross-entropy loss (CE). They have shown that PGD-based adversarial training could improve the adversarial robustness against various adversarial attacks. It marked a milestone in adversarial training methods, and many variants of AT methods use the PGD adversarial attack to optimize the model.

\noindent \textbf{TRADES \cite{pmlr-v97-zhang19p}:} Zhang et al. theoretically identified a trade-off between adversarial robustness and natural accuracy. From the theoretical analysis, they proposed the surrogate loss that improves adversarial robustness. The loss function consists of two terms. The first term aims to maximize natural accuracy with CE loss and the second term encourages the output to be smoothed by minimizing the KL-divergence between the output of clean images and adversarial images.

\noindent \textbf{MART \cite{Wang2020Improving}:} Wang et al. investigated the influence of misclassified and correctly classified examples on adversarial robustness. They found that the adversarial perturbation on misclassified examples has more impact on the adversarial robustness than correctly classified examples. Then, they proposed the surrogate loss function that considers misclassified examples. The loss function consists of Boost Cross Entropy (BCE) function and the misclassified example aware regularization term. The BCE adds the cross-entropy loss and margin loss terms to improve the decision margin of the model. The misclassified example aware regularization term regularizes the model by weighting the misclassified examples.

\noindent \textbf{GAIRAT \cite{zhang2021geometryaware}:} Zhang et al. proposed a geometry-aware instance-reweighted adversarial training method. They argued that each adversarial image has unequal importance to train the model. In other words, a clean image located near the class boundary is less robust, and the corresponding adversarial image should be assigned with a larger weight. Therefore, they proposed a weight function that weights cross-entropy loss according to how robust the input image is. If the image requires a small number of iterations to change the decision during the adversarial perturbation optimization, the weight has a large value.

The aforementioned methods try to improve adversarial robustness by optimizing model parameters. Different from these works, in this paper, we propose a new insight that improves adversarial robustness and natural accuracy by optimizing the external signal. By injecting the optimized booster signal into the input, it makes the input itself to be robust by reducing the input gradient. Also, since the booster signal is separated from the model parameters, we can optimize the booster signal in parallel to any existing AT methods. In other words, once the AT methods improve the natural and robust accuracy by optimizing the model parameters, we can boost them further by optimizing the booster signal collaboratively.

\subsection{Defense: Gradient Masking}
Besides adversarial training methods, many methods for improving adversarial robustness have been proposed. It includes randomization \cite{raff2019barrage,xie2017mitigating,lee2020robust,liu2018towards,raff2019barrage,aprilpyone2021block} and purification \cite{das2017keeping,liu2019feature,naseer2020self,meng2017magnet,song2018pixeldefend,naseer2019local}. In the early research, these research have been widely conducted. However, these approaches degenerate the gradient of the target model and induce gradient masking. As discussed in \cite{athalye2018obfuscated}, defense methods with gradient masking are ineffective under adaptive attack settings constructed using expectation over the transforms or gradient approximation.

Different from these methods, our method aims to be effective under the adaptive attack setting. In other words, even though the external signal is exposed to the adversary, we aim to defend against external signal-aware adversarial attacks and do not suffer from gradient masking.

\section{Motivation and Observation}
The motivation of the proposed method is to improve robustness and natural accuracy by injecting an external signal other than model parameters. To this end, in this section, we define the external signal and describe how to inject the external signal during the training and inference time. Then, through proof-of-concept experiments, we observe the possibility of improving the adversarial robustness by injecting the external signal.

\subsection{External Signal Injection}
We define the external signal as a signal injected into the outside of the input data and call it a \textit{Booster Signal}. In the proposed method, the booster signal is injected into the input image through a signal injection module. Fig. \ref{fig:2} shows the overview of the proposed optimization process for the booster signal and model parameters. As shown in the figure, in the signal injection module, the booster signal with a width of $w$ is injected into the input image. When injecting the signal to the input image, we place the signal to the outside of each input image to satisfy two properties: \textit{i) keep the original image contents} and \textit{ii) increase defensive capability}. Injecting the signal inside the image damages the original contents and might induce the performance to decrease.

Also, most adversarial attack methods could strengthen the attack capability by controlling the magnitude of the perturbations. From a counter-intuitive perspective, we could improve the defensive capability of the signal by controlling the magnitude of the signal. However, since increasing the magnitude of the signal inside the image could hurt the original contents, the magnitude of the signal is limited. Therefore, we place the signal to the outside of the image to increase the defensive capability without limitation of the signal magnitude.

\subsection{Observation of Input Gradient}
In this section, we refer to the gradient of the loss function with respect to the input data as the input gradient for simplicity. The input gradient represents how small changes at each input pixel affect the model prediction. Therefore, the prediction of the input with a large input gradient value is easily changed by a perturbation. On the other hand, even if the perturbation is added to an input with a small input gradient value, the prediction hardly changes. Also, as discussed in Section II. A, most adversarial attack algorithms use the input gradient and it is related to robustness \cite{ross2018improving,chan2020thinks}.

\begin{figure}[!t]
	\centering
	    \includegraphics[width=0.98\linewidth]{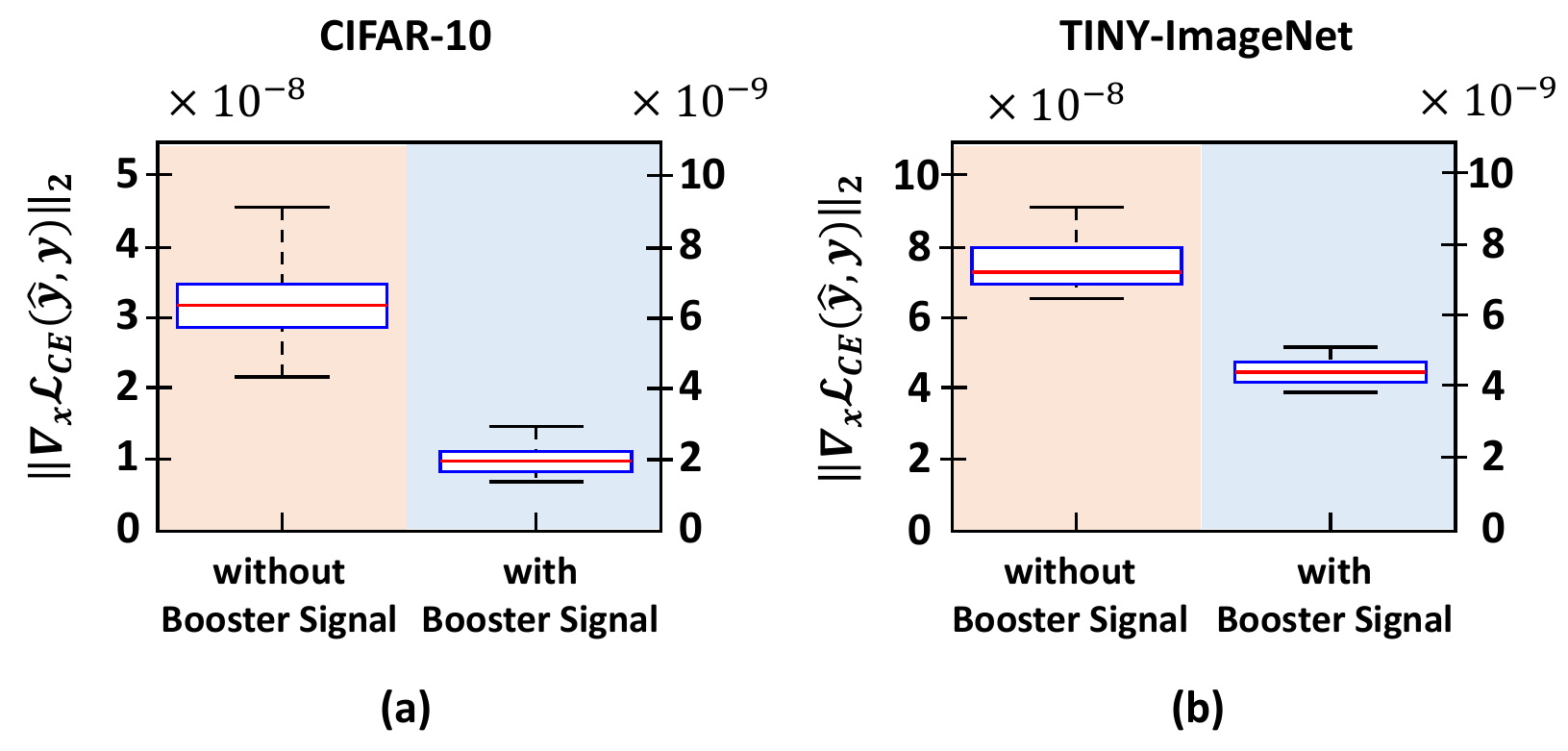}
	\caption{Distribution of L2-norm magnitudes of input gradients. (a) is the input gradient distribution of CIFAR-10 dataset and (b) is the input gradient distribution of the Tiny-ImageNet dataset. Here, the booster signal is individually optimized for each image.} 
\label{fig:3}
\end{figure}

Therefore, we first observe whether injecting the booster signal can reduce the input gradient value. To illustrate this phenomenon, we conduct proof-of-concept experiments on CIFAR-10 and TINY-ImageNet datasets. With given an original image $x$, let $f_{\theta}(x)=p(y|x,\theta)$ be a prediction of the given model, where $\theta$ denotes the parameters of a pretrianed model. We also assume that we are given a suitable loss function $\mathcal{L}$ such as a cross-entropy loss function. The purpose of this section is to find a booster signal ($b$) that reduces the gradient of the loss function with respect to the input data. To this end, we optimize the booster signal according to the following update equation,

\begin{equation}
b^{t+1}=b^{t}-\eta \nabla_{x}\mathcal{L}(f_{\theta}(\mathcal{B}(x,b^{t})),y),
\label{eq:2}
\end{equation}

\noindent where $\mathcal{B}(\cdot)$ injects the booster signal ($b$) to the outside of the image, $b$ denotes the booster signal corresponding to input $x$, $y$ denotes the ground-truth of input, $t$ denotes the number of iterations for optimizing the booster signal, and $\eta$ denotes the constant value that controls the magnitude of update. By using Eq. \ref{eq:2}, we generate booster signals for the entire data set and we statistically analyze the input gradients. Fig. \ref{fig:3} shows the distribution of L2-norm magnitudes of input gradients in two datasets. In Fig. \ref{fig:3}, the booster signals are optimized individually and we apply them to corresponding images. As shown in the figure, injecting the booster signal reduces the L2-norm magnitudes. The observation can be interpreted that injecting the booster signal can make the input to be robust against adversarial attacks.

\noindent \textbf{Challenge:} Although we have verified the possibility of improving the adversarial robustness through the booster signal, it is hard to optimize during the inference. Since the ground-truth label ($y$) is unknown during the inference, it is hard to implement Eq. \ref{eq:2}. Therefore, in the proposed method, instead of optimizing booster signals for every input data, we try to optimize a single booster signal that reduces the expectation of the input gradients for most images (image-agnostic booster signal). In the following section, we describe how to optimize the image-agnostic booster signal and optimize it in cooperation with existing AT methods.

\section{Proposed Booster Defense}

As seen in Fig. \ref{fig:2}, in the proposed framework, we optimize the booster signal and model parameters collaboratively over 4 steps. Red lines and square boxes denote the optimizing model parameters and booster signal respectively. Each step has a signal injection module and a classifier. As shown in Fig. \ref{fig:2}, in the signal injection module, we inject an external signal to the outside of the image set. Then, the signal injected image set is fed into the classifier and computes the loss function ($\mathcal{L}_{model}$ and $\mathcal{L}_{signal}$). For optimizing model parameters, we use $\mathcal{L}_{model}$ and optimize the booster signal with $\mathcal{L}_{signal}$. After the optimization, we used the optimized booster signal and classifier for inference.



\subsection{Robustify Classifier with Adversarial Training}
Fig. \ref{fig:2} (a) describes the first step. In the first step, we train the classifier through adversarial training. In  this  case,  we  train  the model by injecting the null signal ($b_{null}$) at the position where the booster signal will be injected. Then, we train the classifier by minimizing $\mathcal{L}_{model}$, where $\mathcal{L}_{model}$ is the existing adversarial training loss function. For example, in the case of MART, we use the misclassified example-aware loss proposed in \cite{Wang2020Improving}. Through the first step, the classifier could achieve the adversarial robustness that existing methods have achieved.

\subsection{Booster Signal Optimization}
In this section, we define the formulation of the booster signal and introduce how to optimize it. We consider a booster signal that could \textit{i) improve natural accuracy} and \textit{ii) improve adversarial robustness}. The basic intuition behind our method is that we can optimize a booster signal that could transform the input to be correctly classified. Considering that adding adversarial perturbation to input space could transfer the correctly classified example to be misclassified by maximizing the input gradient. Likewise, injecting a well-optimized signal into the input image could transfer the misclassified example to be correctly classified and reduce the input gradient. Therefore, the problem can be defined as follows,
\begin{equation}
    \underset{b}{\textrm{argmin}}\mathcal{L}_{signal}(f_\theta(\mathcal{B}(x,b)),y).\ 
    \label{eq:3}
\end{equation}
\begin{equation}
    \begin{aligned}
            & \underset{b}{\textrm{argmin}}\mathcal{L}_{signal}(f_{\theta}(\mathcal{B}(x,b)+p^{adv}),y), \\ 
            \textrm{where}& \quad  p^{adv}=\underset{||p||<\epsilon}{\textrm{argmax}}\mathcal{L}_{adv}(f_{\theta}(\mathcal{B}(x,b)+p),y),
    \end{aligned}
    \label{eq:4}
\end{equation}

\noindent where $\mathcal{L}_{signal}$ denotes the cross-entropy loss function for optimizing booster signal. Solving Eq. \ref{eq:3} could be interpreted that by injecting booster signal to the input image ($\mathcal{B}(x,b)$), it makes the classifier predict the correct class. Also, Eq. \ref{eq:4} describes that the booster signal is optimized to defend against adversarial perturbations by countering adversarial attacks. The Eq. \ref{eq:4} is solved in recursion. However, as we discussed in Section III. B, we cannot simply solve the problem during the inference since the ground truth $y$ is unknown at inference time. Therefore, we aim to optimize an image-agnostic booster signal that could be applied to any input image. In the following section, we describe how to optimize the booster signal.

\subsubsection{Standard Optimization for Booster Signal}
Fig. \ref{fig:2} (b) shows the visual explanation of the standard optimization for natural accuracy. Let $X=\{x_1,x_2,...,x_m\}$ be a subset of images sampled from the training data distribution $\mu$. Specifically, we randomly sample $m$ number of images for one image subset $X$. Then, we generate $n/m$ number of image subsets, where $n$ denotes the number of total images in the training image set. After generating image subsets, we seek the image-agnostic booster signal ($b$) that makes the prediction to be correct. Therefore, Eq. \ref{eq:3} is transformed as follows:
\begin{equation}
    {\underset{b}{\textrm{argmin}}}\mathcal{L}_{signal}(f_\theta(\mathcal{B}(X,b)),Y),
    \label{eq:5}
\end{equation}
where $Y=\{y_1,y_2,...,y_m\}$ denotes the set of ground truth of input image set $X$ . The optimization process seeks an image-agnostic signal that correctly classifies the data points in $X$. To specify, the booster signal is iteratively updated according to the following equation,
\begin{equation}
\begin{aligned}
    b^{t+1}=b^t-\nabla_b\mathcal{L}_{signal}(f_\theta(\mathcal{B}(X,b^{t})),Y),\\
\textrm{where} \quad \mathcal{L}_{signal}(\hat{Y},Y)=\ \mathbb{E}_{X\sim\mu}[\textrm{CE}(\hat{Y},Y)],
\end{aligned}
\label{eq:6}
\end{equation}
where $\hat{Y}$ denotes the predictions of classifier and CE($\cdot$) denotes the cross-entropy loss. We have chosen a greedy algorithm to optimize $b$. The algorithm iteratively runs over all data points of $X$. At each iteration, we compute the $\nabla_{b}$ to correctly classify the booster signal injected input $\mathcal{B}(X,b)$. The optimization process terminates until $K$-\textit{th} iteration. Through optimization, 
the booster signal makes the expectation of the input gradient to be reduced and clean images could be classified correctly for the data points in $X$. After the optimization, it is repeated for all image subsets. Then, the optimized booster signal could be general and flexible enough to be applicable to any images.

\begin{figure}[!t]
	\centering
	    \includegraphics[width=0.85\linewidth]{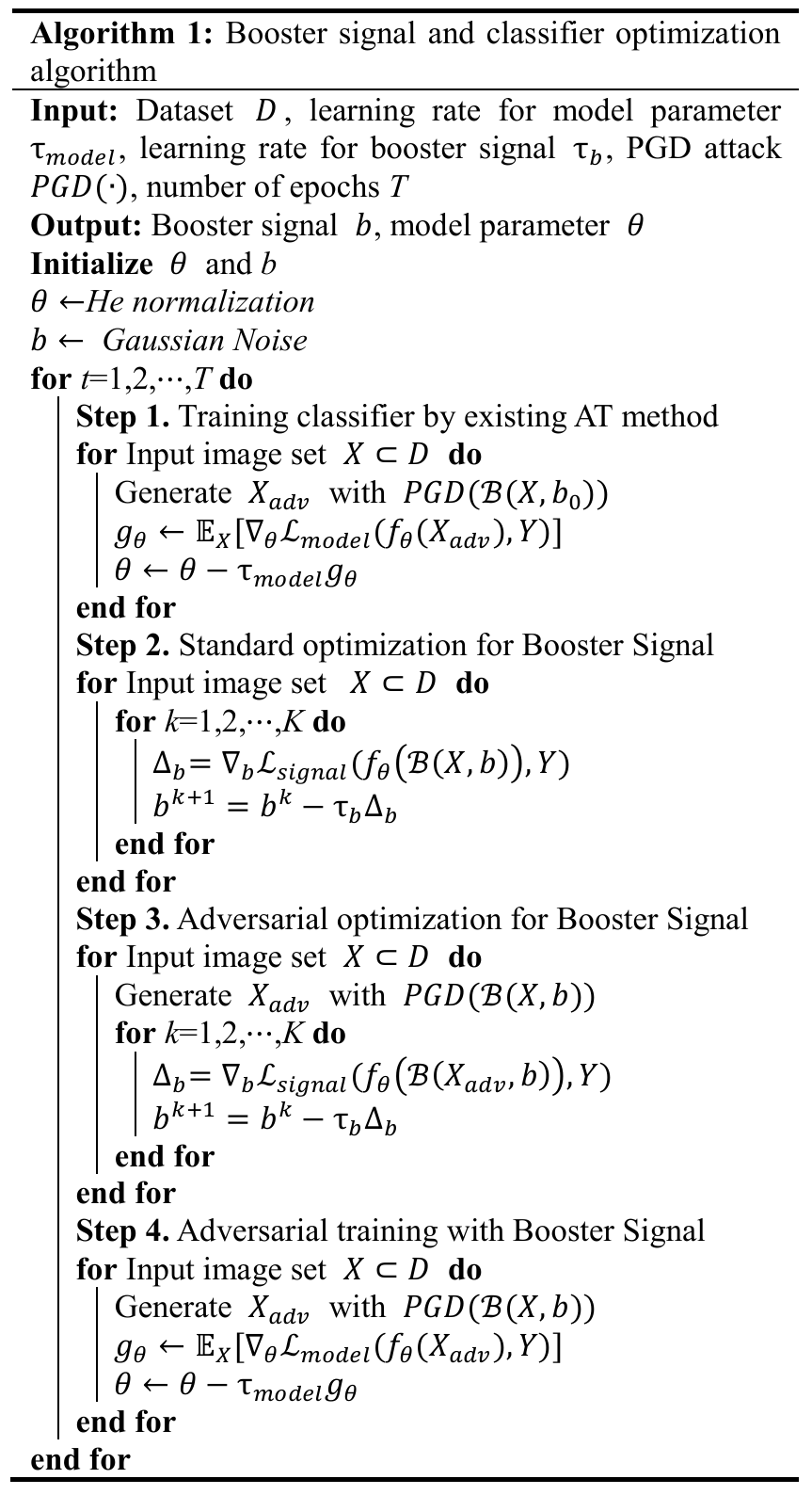}
	    \vspace{-0.5cm}
\end{figure}

\subsubsection{Adversarial Optimization for Booster Signal}
To improve the adversarial robustness, we conduct adversarial optimization. Fig. \ref{fig:2} (c) shows the visual explanation of adversarial optimization for adversarial robustness. The basic intuition is similar to standard optimization. We optimize the booster signal by minimizing the following objective:

\begin{equation}
    \begin{aligned}
        & {\underset{b}{\textrm{argmin}}}{\mathcal{L}_{signal}(f_\theta(\mathcal{B}(X,b)+P),}Y), \\
        \textrm{where} & \quad  p_i = \underset{||p_i||<\epsilon}{\textrm{argmax}}\mathcal{L}_{adv}(f_\theta(\mathcal{B}(x_i,b)+p_i),y_i),
    \end{aligned}
    \label{eq:7}
    \vspace{0.3cm}
\end{equation}

\noindent where $P=\{p^{adv}_1,p^{adv}_2,...,p^{adv}_m\}$ denotes the set of adversarial perturbation that attacks corresponding input $\mathcal{B}(x_i,b)$. Eq. \ref{eq:7} optimizes a booster signal to minimize the difference between the prediction of the adversarial example set ($f_\theta(\mathcal{B}(X,b)+P)$) and set of ground-truth by using cross-entropy loss. Also, the Eq. \ref{eq:7} is solved in recursion. Following Eq. \ref{eq:7}, the adversarial perturbation is optimized to attack the booster signal injected input, then the booster signal is optimized to counter the adversarial perturbation. Therefore, we iteratively update the adversarial perturbation and the booster signal in an adversarial manner. The equation can be written as follows,
\begin{equation}
    p_i^{k+1}=p_i^k+\nabla_p\mathcal{L}_{adv}(f_\theta(\mathcal{B}(x_i,b)+p_i^k),y_i),
\label{eq:8}
\end{equation}
\begin{equation}
    b^{t+1}=b^t-\nabla_b\mathcal{L}_{signal}(f_\theta(\mathcal{B}(X,b^{t})+P),Y),
    \label{eq:9}
\end{equation}
where $k$ is an iteration step for generating PGD adversarial perturbation. During the optimization process, the individual adversarial perturbations are optimized to attack individual input $\mathcal{B}(x_i,b)$ through the PGD attack method, and the booster signal is optimized to defend input $\mathcal{B}(X,b)+P$ by optimizing Eq. \ref{eq:9}. Through the optimization process, the booster signal has the ability to defend against adversarial attacks that attacks the booster signal. In other words, even though the attacker knows the existence of the booster signal, we could defend against white-box attacks. Also, reducing the expectation of gradient of adversarial input makes the signal injected input itself to be robust against adversarial attacks.

\subsection{Adversarial Training with Booster Signal}
After we optimize the booster signal, we train the classifier with booster signal injected inputs. In the fourth step, we use the booster signal optimized in step 3. Since injecting the booster signal changes the data distribution, the classifier further to be trained to fit the changed distribution. Therefore, the adversarial perturbation is generated on the booster signal injected image ($\mathcal{B}(x,b)$) by using PGD attack algorithm. Then, with the adversarial perturbation, we train the classifier by minimizing $\mathcal{L}_{model}$ as previous adversarial training approaches did. We summarize the whole optimization process in Algorithm 1. As seen in the algorithm, in each step, parameters are optimized for entire sub-image sets, then proceed to the next step. We conduct this process for every step.

\section{Experimental Results}
\subsection{Experiment Setting}
\noindent \textbf{Dataset:} We conduct experiments to verify the effectiveness of our proposed booster signal defense framework on three benchmark datasets (CIFAR-10 \cite{krizhevsky2009learning}, Tiny-ImageNet \cite{le2015tiny}, and ImageNet \cite{russakovsky2015imagenet}). The spatial resolutions are $32\times32$ for CIFAR-10 and $64\times64$ for Tiny-ImageNet. In the case of the ImageNet dataset, we cropped and resized an image with a size of 288 following the protocol of \cite{Wong2020Fast}. To optimize the booster signal, we set the width of the booster signal to $w=5,10,40$ for CIFAR-10, Tiny-ImageNet, and ImageNet datasets respectively.

\begin{table*}[!t]
\caption{Adversarial robust accuracy and natural accuracy (Clean) on CIFAR-10 dataset under white-box attack setting with WideResNet-28-10.}
\centering
\resizebox{0.6\linewidth}{!}{
\begin{tabular}{ccccccc}
\specialrule{.15em}{.1em}{.1em}
\multicolumn{2}{c}{\textbf{Method}}                                      & \textbf{Natural} & \textbf{FGSM}  & \textbf{PGD-20} & \textbf{CW}    & \textbf{AutoAttack} \\ \hline
\multicolumn{1}{c}{\multirow{3}{*}{\textbf{Madry}}}  & Base              & 85.59            & 59.38          & 54.21           & 49.19         & 47.93               \\
\multicolumn{1}{c}{}                                 & Ours (w/o signal) & 85.06            & 59.50          & 54.64           & 49.07          & 47.00               \\
\multicolumn{1}{c}{}                                 & Ours              & \textbf{86.69}   & \textbf{62.45} & \textbf{57.50}  & \textbf{51.28} & \textbf{52.32}      \\ \hline
\multicolumn{1}{c}{\multirow{3}{*}{\textbf{TRADES}}}  & Base              & 85.28            & 61.47          & 56.24           & 50.79          & 49.85               \\
\multicolumn{1}{c}{}                                 & Ours (w/o signal) & 85.80            & 63.28          & 56.68           & 51.77          & 50.82               \\
\multicolumn{1}{c}{}                                 & Ours              & \textbf{87.13}   & \textbf{65.24} & \textbf{58.08}  & \textbf{53.50} & \textbf{52.80}      \\ \hline
\multicolumn{1}{c}{\multirow{3}{*}{\textbf{MART}}}   & Base              & 85.71            & 61.54          & 56.23           & 52.41          & 51.01               \\
\multicolumn{1}{c}{}                                 & Ours (w/o signal) & 84.74            & 61.93          & 54.24           & 52.26          & 51.46               \\
\multicolumn{1}{c}{}                                 & Ours              & \textbf{87.29}   & \textbf{64.22} & \textbf{58.33}  & \textbf{54.84} & \textbf{53.95}      \\ \hline
\multicolumn{1}{c}{\multirow{3}{*}{\textbf{GAIRAT}}} & Base              & 84.56            & 62.51          & 57.82           & 44.38          & 40.51               \\
\multicolumn{1}{c}{}                                 & Ours (w/o signal) & 85.61            & 67.80          & 57.17           & 44.00          & 40.12               \\
\multicolumn{1}{c}{}                                 & Ours              & \textbf{87.82}   & \textbf{69.07} & \textbf{59.40}  & \textbf{45.01} & \textbf{42.11}      \\ \hline
\multicolumn{1}{c}{\multirow{3}{*}{\textbf{FAT}}} & Base              & 87.48            & 61.51          & 48.28           & 47.27          & 46.72               \\
\multicolumn{1}{c}{}                                 & Ours (w/o signal) & 85.31            & 65.01          & 48.07           & 46.91          & 46.57               \\
\multicolumn{1}{c}{}                                 & Ours              & \textbf{87.92}   & \textbf{67.31} & \textbf{49.85}  & \textbf{48.79} & \textbf{47.75}      \\ \hline
\multicolumn{1}{c}{\multirow{3}{*}{\textbf{HAT}}} & Base              & 86.85            & 63.08          & 56.75           & 53.92          & 52.52               \\
\multicolumn{1}{c}{}                                 & Ours (w/o signal) & 85.73            & 63.80          & 56.17           & 53.00          & 52.09               \\
\multicolumn{1}{c}{}                                 & Ours              & \textbf{87.95}   & \textbf{65.70} & \textbf{58.40}  & \textbf{55.89} & \textbf{54.55}      \\
\specialrule{.15em}{.1em}{.1em}

\end{tabular}}
\label{tab:1}

\end{table*}
\begin{table*}[!t]
\caption{Adversarial robust accuracy and natural accuracy (Clean) on CIFAR-10 dataset under white-box attack setting with ResNet-18.}
\centering
\resizebox{0.6\linewidth}{!}{%
\begin{tabular}{ccccccc}
\specialrule{.15em}{.1em}{.1em}
\multicolumn{2}{c}{\textbf{Method}}                    & \textbf{Natural} & \textbf{FGSM}  & \textbf{PGD-20} & \textbf{CW}    & \textbf{AutoAttack} \\ \hline
\multirow{3}{*}{\textbf{Madry}}  & Base                & 83.81            & 57.35          & 49.15           & 48.34          & 46.02               \\
                                 & Ours (w/o   signal) & 83.02            & 56.86          & 49.07           & 46.17          & 46.00               \\
                                 & Ours                & \textbf{84.89}   & \textbf{59.27} & \textbf{51.28}  & \textbf{48.45} & \textbf{47.25}      \\ \hline
\multirow{3}{*}{\textbf{TRADES}} & Base                & 83.01            & 59.41          & 53.09           & 48.60          & 48.01               \\
                                 & Ours (w/o   signal) & 83.58            & 60.15          & 53.89           & 48.01          & 48.06               \\
                                 & Ours                & \textbf{84.23}   & \textbf{60.74} & \textbf{55.11}  & \textbf{49.65} & \textbf{49.25}      \\ \hline
\multirow{3}{*}{\textbf{MART}}   & Base                & 82.37            & 58.65          & 54.11           & 48.59          & 47.24               \\
                                 & Ours (w/o   signal) & 81.98            & 59.06          & 54.26           & 49.01          & 47.28               \\
                                 & Ours                & \textbf{84.02}   & \textbf{60.66} & \textbf{55.62}  & \textbf{50.76} & \textbf{49.74}      \\ \hline
\multirow{3}{*}{\textbf{GAIRAT}} & Base                & 82.53            & 59.73          & 55.81           & 42.28          & 38.72               \\
                                 & Ours (w/o   signal) & 82.81            & 59.51          & 56.02           & 42.17          & 38.33               \\
                                 & Ours                & \textbf{84.20}   & \textbf{61.31} & \textbf{56.72}  & \textbf{43.30} & \textbf{39.72}      \\ \hline
\multirow{3}{*}{\textbf{FAT}}    & Base                & 86.42            & 59.35          & 46.17           & 45.51          & 43.91               \\
                                 & Ours (w/o   signal) & 87.01            & 60.15          & 46.23           & 45.81          & 44.51               \\
                                 & Ours                & \textbf{87.43}   & \textbf{61.07} & \textbf{47.37}  & \textbf{46.53} & \textbf{45.54}      \\ \hline
\multirow{3}{*}{\textbf{HAT}}    & Base                & 84.09            & 59.98          & 52.04           & 49.80          & 48.61               \\
                                 & Ours (w/o   signal) & 84.23            & 60.22          & 53.06           & 50.22          & 49.07               \\
                                 & Ours                & \textbf{85.09}   & \textbf{61.20} & \textbf{54.29}  & \textbf{51.33} & \textbf{50.48}      \\ \specialrule{.15em}{.1em}{.1em}
\end{tabular}%
}\label{tab:2}
\end{table*}

\begin{table*}[!t]
\centering
\caption{Adversarial robust accuracy and natural accuracy (Clean) on Tiny-ImageNet dataset under white-box attack setting with WideResNet-28-10.}
\resizebox{0.6\linewidth}{!}{
\begin{tabular}{ccccccc}
\specialrule{.15em}{.1em}{.1em}
\multicolumn{2}{c}{\textbf{Method}}                                      & \textbf{Natural} & \textbf{FGSM}  & \textbf{PGD-20} & \textbf{CW}    & \textbf{AutoAttack} \\ \hline
\multicolumn{1}{c}{\multirow{3}{*}{\textbf{Madry}}}  & Base              & 48.60            & 25.14          & 23.01           & 20.05          & 18.76               \\
\multicolumn{1}{c}{}                                 & Ours (w/o signal) & 48.40            & 25.84          & 23.31           & 20.72          & 18.16                \\
\multicolumn{1}{c}{}                                 & Ours              & \textbf{50.68}   & \textbf{27.19} & \textbf{25.32}  & \textbf{23.16} & \textbf{19.16}      \\ \hline
\multicolumn{1}{c}{\multirow{3}{*}{\textbf{TRADES}}}  & Base              & 50.60            & 26.83          & 25.19           & 21.99          & 19.05                \\
\multicolumn{1}{c}{}                                 & Ours (w/o signal) & 50.84            & 27.37          & 25.90           & 22.35          & 19.67               \\
\multicolumn{1}{c}{}                                 & Ours              & \textbf{52.07}   & \textbf{29.76} & \textbf{28.08}  & \textbf{24.33} & \textbf{21.44}      \\ \hline
\multicolumn{1}{c}{\multirow{3}{*}{\textbf{MART}}}   & Base              & 50.43            & 28.26          & 26.17           & 23.47           & 20.40               \\
\multicolumn{1}{c}{}                                 & Ours (w/o signal) & 50.31            & 29.31          & 27.67           & 23.00          & 21.83               \\
\multicolumn{1}{c}{}                                 & Ours              & \textbf{52.72}   & \textbf{32.05} & \textbf{29.47}  & \textbf{25.23} & \textbf{23.31}      \\ \hline
\multicolumn{1}{c}{\multirow{3}{*}{\textbf{GAIRAT}}} & Base              & 51.16            & 27.10          & 26.40           & 19.47          & 17.01               \\
\multicolumn{1}{c}{}                                 & Ours (w/o signal) & 50.81            & 28.05          & 26.72           & 19.35          & 18.03               \\
\multicolumn{1}{c}{}                                 & Ours              & \textbf{53.56}   & \textbf{30.05} & \textbf{29.45}  & \textbf{21.62} & \textbf{20.86}      \\ \hline
\multicolumn{1}{c}{\multirow{3}{*}{\textbf{FAT}}} & Base              & 51.48            & 27.15          & 20.81           & 19.19          & 18.33               \\
\multicolumn{1}{c}{}                                 & Ours (w/o signal) & 51.31            & 27.21          & 21.07           & 19.84          & 18.77               \\
\multicolumn{1}{c}{}                                 & Ours              & \textbf{53.92}   & \textbf{29.64} & \textbf{23.15}  & \textbf{21.79} & \textbf{20.65}      \\ \hline
\multicolumn{1}{c}{\multirow{3}{*}{\textbf{HAT}}} & Base              & 52.65            & 27.80          & 26.75           & 23.52          & 20.03               \\
\multicolumn{1}{c}{}                                 & Ours (w/o signal) & 51.97            & 27.08          & 26.17           & 23.00          & 20.09               \\
\multicolumn{1}{c}{}                                 & Ours              & \textbf{53.15}   & \textbf{29.70} & \textbf{27.40}  & \textbf{24.99} & \textbf{22.35}      \\
\specialrule{.15em}{.1em}{.1em}
\end{tabular}}
\label{tab:3}

\end{table*}

\begin{table*}[!t]
\centering
\caption{Adversarial robust accuracy and natural accuracy (clean) on ImageNet dataset under white-box attack setting.}
\resizebox{0.66\linewidth}{!}{%
\begin{tabular}{ccccccc}
\specialrule{.15em}{.1em}{.1em}
\multicolumn{2}{c}{\textbf{Method}}                              & \textbf{Natural} & \textbf{FGSM}  & \textbf{PGD}   & \textbf{CW}    & \textbf{AutoAttack} \\ \hline
\multirow{3}{*}{\textbf{Fast AT}} & Base              & 55.45            & 40.72          & 31.24          & 26.45          & 23.84               \\
                                             & Ours (w/o signal) & 55.01            & 40.52          & 32.65          & 27.04          & 23.01               \\
                                             & Ours              & \textbf{56.68}   & \textbf{42.03} & \textbf{33.47} & \textbf{28.67} & \textbf{25.50}     \\
\specialrule{.15em}{.1em}{.1em}
\end{tabular}%
}\label{tab:4}
\end{table*}

\noindent \textbf{Attack Methods:} In the proposed method, we focus on defending against adversarial attacks that imperceptibly manipulate the input image. To evaluate the defensive performance of the proposed defense framework on such attacks, we apply four adversarial attack methods widely used to evaluate defensive performance (FGSM \cite{goodfellow2014explaining}, PGD \cite{madry2017towards}, CW \cite{carlini2017towards}, and AutoAttack \cite{croce2020reliable}). These methods attack the image by adding small and imperceptible noise to the input image. In the experiment, we set the perturbation budget $\epsilon=8/255$ for both datasets. For the PGD adversarial attack, we generate adversarial perturbation with 20 optimization steps with the step size $\epsilon/10$. For the CW attack, we use L2-norm bounded attacks with 200 iterations and use ADAM optimizer. Also, in the case of AutoAttack, we use three attack methods (APGD-CE, APGD-DLR \cite{croce2020reliable}, FAB \cite{croce2020minimally}, and Square Attack \cite{andriushchenko2020square}). For FAB attack hyper-parameters, we optimize the perturbation with 100 iterations and 5 random restarts. In the case of the Square attack, we fed 5000 queries for the black-box attack.






\noindent \textbf{Defense Baselines:}
We apply our booster defense framework to six recently proposed state-of-the-art adversarial training methods (Madry \cite{madry2017towards}, MART \cite{Wang2020Improving}\footnote[1]{\href{https://github.com/YisenWang/MART}{https://github.com/YisenWang/MART}}, TRADES \cite{pmlr-v97-zhang19p}\footnote[2]{\href{https://github.com/yaodongyu/TRADES}{https://github.com/yaodongyu/TRADES}}, GAIRAT \cite{zhang2021geometryaware}\footnote[3]{\href{https://github.com/zjfheart/Geometry-aware-Instance-reweighted-Adversarial-Training}{https://github.com/zjfheart/Geometry-aware-Instance-reweighted-Adversarial-Training}} FAT \cite{zhang2020attacks}\footnote[4]{\href{https://github.com/zjfheart/Friendly-Adversarial-Training}{https://github.com/zjfheart/Friendly-Adversarial-Training}} and HAT \cite{rade2021helper}\footnote[5]{\href{https://github.com/imrahulr/hat}{https://github.com/imrahulr/hat}}). The Madry, MART, TRADES, and GAIRAT are widely used AT methods that improve the adversarial robustness. FAT and HAT are recently proposed AT methods that handle the problem of trade-offs. For the evaluation, we use the WideResNet-28-10 network \cite{zagoruyko2016wide} and ResNet-18 as classifiers. We use them as base networks and set the batch size as 256. To generate PGD adversarial perturbation, we set the epsilon budget as $\epsilon=8/255$, step size $\alpha=\epsilon/4$ with 7 iterations. For both datasets, the model is trained using the SGD algorithm. In the case of the ImageNet dataset, since it requires extremely large computation costs for training the model with existing methods, we adapt a fast adversarial training strategy (Fast AT) \cite{Wong2020Fast}\footnote[6]{\url{https://github.com/locuslab/fast_adversarial}} on ResNet-50. We train the model to be robust at $\epsilon=4/255$ and the batch size is set as 128.

\begin{table}[!t]
\centering
\caption{Black-box attack evaluation on CIFAR-10 dataset. The perturbation is generated on WideResNet-34-10.}

\resizebox{0.98\linewidth}{!}{
\begin{tabular}{cccccc}
\specialrule{.15em}{.1em}{.1em}
\multicolumn{2}{c}{\textbf{Method}}                         & \textbf{FGSM}  & \textbf{PGD-20} & \textbf{CW}    & \textbf{AutoAttack} \\ \hline
\multicolumn{1}{c}{\multirow{2}{*}{\textbf{Madry}}}  & Base & 81.37          & 82.41           & 83.02          & 82.41               \\
\multicolumn{1}{c}{}                                 & Ours & \textbf{82.12} & \textbf{83.89}  & \textbf{84.21} & \textbf{83.91}      \\ \hline
\multicolumn{1}{c}{\multirow{2}{*}{\textbf{TRADES}}}  & Base & 82.29          & 83.01           & 83.16          & 82.98               \\
\multicolumn{1}{c}{}                                 & Ours & \textbf{83.21} & \textbf{84.39}  & \textbf{84.84} & \textbf{84.72}      \\ \hline
\multicolumn{1}{c}{\multirow{2}{*}{\textbf{MART}}}   & Base & 81.76          & 82.56           & 83.09          & 82.59               \\
\multicolumn{1}{c}{}                                 & Ours & \textbf{83.37} & \textbf{84.00}  & \textbf{84.27} & \textbf{84.21}      \\ \hline
\multicolumn{1}{c}{\multirow{2}{*}{\textbf{GAIRAT}}} & Base & 81.17          & 82.15           & 82.62          & 82.19               \\
\multicolumn{1}{c}{}                                 & Ours & \textbf{84.66} & \textbf{85.06}  & \textbf{85.42} & \textbf{85.11}      \\ \hline
\multicolumn{1}{c}{\multirow{2}{*}{\textbf{FAT}}} & Base & 82.01          & 83.31           & 83.77          & 83.90               \\
\multicolumn{1}{c}{}                                 & Ours & \textbf{84.59} & \textbf{85.01}  & \textbf{85.34} & \textbf{86.01}      \\ \hline
\multicolumn{1}{c}{\multirow{2}{*}{\textbf{HAT}}} & Base & 82.33          & 83.03           & 83.62          & 83.81               \\
\multicolumn{1}{c}{}                                 & Ours & \textbf{84.16} & \textbf{85.60}  & \textbf{85.42} & \textbf{85.88}      \\ 
\specialrule{.15em}{.1em}{.1em}
\end{tabular}}

\label{tab:5}

\end{table}


\begin{table}[!t]
\centering
\caption{Black-box attack evaluation on Tiny-ImageNet dataset. The perturbation is generated on WideResNet-34-10.}
\resizebox{0.98\linewidth}{!}{
\begin{tabular}{cccccc}
\specialrule{.15em}{.1em}{.1em}
\multicolumn{2}{c}{\textbf{Method}}                         & \textbf{FGSM}  & \textbf{PGD-20} & \textbf{CW}    & \textbf{AutoAttack} \\ \hline
\multicolumn{1}{c}{\multirow{2}{*}{\textbf{Madry}}}  & Base & 45.42          & 46.09           & 47.36          & 46.27               \\
\multicolumn{1}{c}{}                                 & Ours & \textbf{47.45} & \textbf{47.99}  & \textbf{48.35} & \textbf{48.23}      \\ \hline
\multicolumn{1}{c}{\multirow{2}{*}{\textbf{TRADESS}}}  & Base & 46.30          & 47.23           & 48.15          & 48.52               \\
\multicolumn{1}{c}{}                                 & Ours & \textbf{48.45} & \textbf{49.03}  & \textbf{49.23} & \textbf{49.98}      \\ \hline
\multicolumn{1}{c}{\multirow{2}{*}{\textbf{MART}}}   & Base & 47.40          & 48.14           & 48.27          & 48.40               \\
\multicolumn{1}{c}{}                                 & Ours & \textbf{48.30} & \textbf{49.15}  & \textbf{49.61} & \textbf{49.50}      \\ \hline
\multicolumn{1}{c}{\multirow{2}{*}{\textbf{GAIRAT}}} & Base & 49.01          & 49.73           & 50.09          & 49.91               \\
\multicolumn{1}{c}{}                                 & Ours & \textbf{49.71} & \textbf{50.22}  & \textbf{51.84} & \textbf{51.02}      \\ \hline
\multicolumn{1}{c}{\multirow{2}{*}{\textbf{FAT}}}   & Base & 49.00          & 49.51           & 49.87          & 49.99               \\
\multicolumn{1}{c}{}                                 & Ours & \textbf{49.68} & \textbf{50.03}  & \textbf{50.51} & \textbf{50.70}      \\ \hline
\multicolumn{1}{c}{\multirow{2}{*}{\textbf{HAT}}} & Base & 48.91          & 49.47           & 49.88          & 50.01               \\
\multicolumn{1}{c}{}                                 & Ours & \textbf{49.71} & \textbf{49.93}  & \textbf{50.64} & \textbf{50.98}      \\
\specialrule{.15em}{.1em}{.1em}
\end{tabular}}

\label{tab:6}

\end{table}

\subsection{Adversarial Robustness Evaluation}

\subsubsection{White-box Evaluation}
To evaluate the proposed method, we optimize the booster signal with six recently proposed AT methods. Table \ref{tab:1} shows the natural accuracy and robust accuracy on the CIFAR-10 dataset with WideResNet-28-10, where Base denotes the results of implementing existing AT methods and Ours denotes the results of applying our proposed defense framework to existing AT methods. To verify the effectiveness of the proposed method, we conduct the experiments under a white-box attack setting. Note that the adversarial perturbation is generated to attack the booster signal injected images. As shown in the table, injecting the booster signal could improve the adversarial robustness regardless of the attack methods. Also, in the case of natural accuracy, the booster signal could improve the natural accuracy. In the case of the w/o Signal, it is the result of using only the classifier without using the booster signal (Using $\mathcal{B}(x,b_{null})$ or $\mathcal{B}(x,b_{null})+p$ as input). It shows similar robustness compared to the Base method. Then, our proposed method can guarantee similar results to the existing AT methods and boost both natural and robust accuracies. Similar results are shown in Table \ref{tab:2}, where the backbone model is ResNet-18. As shown in the table, the proposed method still improves both clean accuracy and robust accuracy. It can be interpreted that optimizing the booster signal is general and flexible enough to be adopted on any existing adversarial training method regardless of model types and sizes. Therefore, once an adversarial training method that optimizes the model parameter is proposed, our proposed method can boost the robustness and natural accuracy of that AT model by optimizing the booster signal.

Table \ref{tab:3} shows the natural accuracy and robust accuracy on Tiny-ImageNet. As shown in the table, our proposed method is still effective on Tiny-ImageNet. Furthermore, we conduct the experiment to verify the effectiveness of the proposed method at larger image sizes. To this end, we use the ImageNet dataset, and the result is shown in Table \ref{tab:4}. As shown in the table, our proposed method is still effective with larger size of images. In the case of natural accuracy, by adding the booster signal, the performance is increased by 1.23\%. Also, the robust accuracy against AutoAttack improves by 1.7\%. 

\begin{table}[!t]
\centering
\caption{Black-Box attack evaluation on CIFAR-10 dataset. The perturbation is generated on VGG-16 network.}

\resizebox{0.98\linewidth}{!}{
\begin{tabular}{cccccc}
\specialrule{.15em}{.1em}{.1em}
\multicolumn{2}{c}{\textbf{Method}}     & \textbf{FGSM}  & \textbf{PGD-20} & \textbf{CW}    & \textbf{AutoAttack} \\ \hline
\multirow{2}{*}{\textbf{Madry}}  & Base & 81.27          & 82.82           & 83.57          & 82.17               \\
                                 & Ours & \textbf{82.13} & \textbf{83.54}  & \textbf{84.24} & \textbf{83.01}      \\ \hline
\multirow{2}{*}{\textbf{TRADES}} & Base & 82.48          & 83.36           & 83.72          & 83.80               \\
                                 & Ours & \textbf{83.01} & \textbf{84.57}  & \textbf{84.01} & \textbf{83.92}      \\ \hline
\multirow{2}{*}{\textbf{MART}}   & Base & 82.21          & 82.79           & 83.10          & 83.21               \\
                                 & Ours & \textbf{83.01} & \textbf{84.41}  & \textbf{84.75} & \textbf{84.88}      \\ \hline
\multirow{2}{*}{\textbf{GAIRAT}} & Base & 82.17          & 82.38           & 84.01          & 84.20               \\
                                 & Ours & \textbf{84.21} & \textbf{83.01}  & \textbf{85.25} & \textbf{85.50}      \\ \hline
\multirow{2}{*}{\textbf{FAT}}    & Base & 82.52          & 83.15           & 84.14          & 84.33               \\
                                 & Ours & \textbf{84.16} & \textbf{85.81}  & \textbf{85.27} & \textbf{86.58}      \\ \hline
\multirow{2}{*}{\textbf{HAT}}    & Base & 82.76          & 83.13           & 83.91          & 83.27               \\
                                 & Ours & \textbf{84.34} & \textbf{85.68}  & \textbf{85.67} & \textbf{85.76}      \\
\specialrule{.15em}{.1em}{.1em}
\end{tabular}%
} \label{tab:7}
\end{table}

\begin{table}[!t]
\centering
\caption{Black-Box attack evaluation on Tiny-ImageNet dataset. The perturbation is generated on VGG-16 network.}

\resizebox{0.98\linewidth}{!}{
\begin{tabular}{cccccc}
\specialrule{.15em}{.1em}{.1em}
\multicolumn{2}{c}{\textbf{Method}}     & \textbf{FGSM}  & \textbf{PGD-20} & \textbf{CW}    & \textbf{AutoAttack} \\ \hline
\multirow{2}{*}{\textbf{Madry}}  & Base & 46.01          & 46.73           & 47.82          & 46.91               \\
                                 & Ours & \textbf{47.84} & \textbf{48.34}  & \textbf{48.56} & \textbf{47.13}      \\ \hline
\multirow{2}{*}{\textbf{TRADES}} & Base & 46.82          & 47.72           & 48.86          & 48.73               \\
                                 & Ours & \textbf{48.92} & \textbf{49.50}  & \textbf{49.17} & \textbf{49.78}      \\ \hline
\multirow{2}{*}{\textbf{MART}}   & Base & 47.77          & 48.34           & 48.67          & 48.56               \\
                                 & Ours & \textbf{84.51} & \textbf{49.88}  & \textbf{50.01} & \textbf{50.10}      \\ \hline
\multirow{2}{*}{\textbf{GAIRAT}} & Base & 49.65          & 49.83           & 50.21          & 49.14               \\
                                 & Ours & \textbf{50.14} & \textbf{50.64}  & \textbf{51.31} & \textbf{50.70}      \\ \hline
\multirow{2}{*}{\textbf{FAT}}    & Base & 49.71          & 49.83           & 50.20          & 50.36               \\
                                 & Ours & \textbf{49.98} & \textbf{50.31}  & \textbf{50.87} & \textbf{51.01}      \\ \hline
\multirow{2}{*}{\textbf{HAT}}    & Base & 49.84          & 49.52           & 50.17          & 50.33               \\
                                 & Ours & \textbf{50.15} & \textbf{50.34}  & \textbf{50.61} & \textbf{51.21}      \\
\specialrule{.15em}{.1em}{.1em}
\end{tabular}%
}\label{tab:8}
\end{table}

\subsubsection{Black-box Evaluation}
Black-box attacks are crafted from clean images by attacking an unknown model. To verify the robustness of the proposed method under the black-box attack settings, we separately train WideResNet-34-10 and VGG-16 then generate adversarial perturbations by FGSM, PGD-20, CW, and AutoAttack. The black-box attack results are shown in Table \ref{tab:5},\ref{tab:6},\ref{tab:7}, and \ref{tab:8}. Table \ref{tab:5} and \ref{tab:6} show the black-box results where the adversarial perturbations are generated on WideResNet-34-10. Then, Table \ref{tab:5} shows the black-box results on the CIFAR-10 dataset, and Table \ref{tab:6} shows the black-box attack results on Tiny-ImageNet. As seen in the tables, our method could boost the adversarial robustness of existing AT methods. Compared with the white-box results, we achieve better robustness against black-box attacks, and it shows close to the natural accuracy. Furthermore, we craft adversarial perturbation from the dissimilar architecture (VGG-16). Table \ref{tab:7} and \ref{tab:8} show the black-box results where the adversarial perturbations are generated on VGG-16. Similarly, as shown in the table, although the adversarial perturbation is crafted from the dissimilar architecture, our proposed method is still effective under black-box attacks. The results suggest that the proposed booster defense is a practical defense scenario whether the model is exposed to the attacker or not.

\begin{figure*}[t]
	\centering
	    \includegraphics[width=0.95\linewidth]{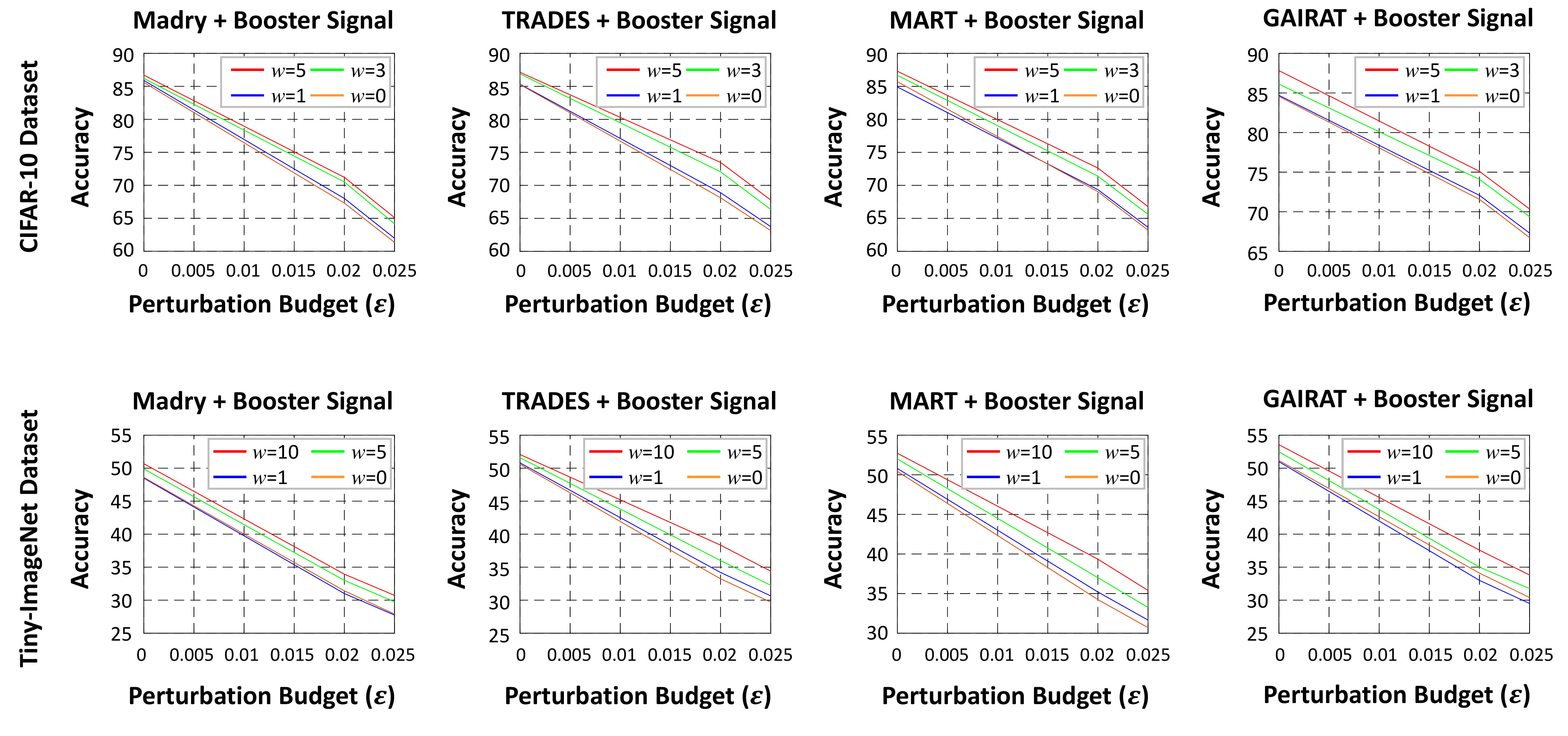}
	\vspace{-0.3cm}
	\caption{The \textit{accuracy vs. perturbation budget} curves according to the signal width ($w$) on CIFAR-10 and Tiny-ImageNet against PGD-20 attack. Note `$w$=0' means baseline adversarial training method.} 

	\label{fig:4}
\vspace{-0.5cm}
\end{figure*}


\subsection{Robustness Evaluation According to Signal Width}
\label{sec:4.3}
In this section, we analyze the effect of the booster signal width $w$. For the analysis, we change the signal width to $w=0,1,3,5$ for CIFAR-10 and $w=0,1,5,10$ for Tiny-ImageNet. Fig. \ref{fig:4} describes robust accuracy vs. perturbation budget curves on CIFAR-10 and Tiny-ImageNet datasets against PGD-20 attack. As shown in the figure, when the signal width is 1 ($w=1$), the adversarial robustness is similar to baseline results ($w=0$). However, as shown in the figure, the robustness increases as the signal width increases. It means that we can increase the defense capacity by extending the signal width. 

\textbf{Discussion:} In this section, we verify that increasing the width of the
booster signal can be helpful to increase the adversarial robustness.
However, if the width of the booster signal is increased, the computation cost for inference increases. Therefore, it is necessary to maximize the defensive capability by using a booster signal of an appropriate width in consideration of the computation cost trade-off. For future work, it would be interesting to design an effective objective function for $\mathcal{L}_{signal}$ to release the limitation.


\begin{table}[t]
\centering
\caption{Comparison of existing model-parameter agnostic defense methods on CIFAR-10 dataset. We use pretrained model trained by Madry \cite{madry2017towards}.}
\resizebox{0.98\linewidth}{!}{
\begin{tabular}{ccccc}
\specialrule{.15em}{.1em}{.1em}
\textbf{\begin{tabular}[c]{@{}c@{}}Defense\\ (Madry)\end{tabular}} & \textbf{Natural} & \textbf{PGD-20} & \textbf{CW}    & \textbf{AutoAttack} \\ \hline
{JPEG \cite{das2017keeping}}                                                      & 81.75               & {52.39}           & 46.71          & 52.65               \\
{FS \cite{xu2017feature}}                                                        & {81.96}               & 54.6            & 47.14          & 53.28               \\
{FD \cite{liu2019feature}}                                                        & 72.25               & 54.3            & 48.9           & 48.62               \\
{TVM \cite{guo2018countering}}                                                       & 69.6               & 37.1            & 29.39          & 29.09               \\
{Reverse \cite{Mao_2021_ICCV}}                                                   & 78.95               & {56.39}           & {50.01} & {53.67}               \\ \hline
{Ours}                                                      & \textbf{86.69}            & \textbf{57.50}   & \textbf{51.28}          & \textbf{54.32}      \\ \specialrule{.15em}{.1em}{.1em}
\end{tabular}}

\label{tab:9}

\end{table}

\begin{table}[t]
\centering
\caption{Runtime comparison (ms) with existing model-parameter agnostic defense methods. * For the Reverse attack, since it is impossible to run on a single GPU, we use 4 multi-GPUs to run Reverse defense.}
\begin{tabular}{cc}
\specialrule{.15em}{.1em}{.1em}
\textbf{\begin{tabular}[c]{@{}c@{}}Defense\\ (Madry)\end{tabular}} & \textbf{Runtime (ms)} \\ \hline
\textbf{No Defense}                                                & 22.83                 \\
\textbf{JPEG}                                                      & 65.06                 \\
\textbf{FS}                                                        & 27.91                 \\
\textbf{FD}                                                        & 39.61                 \\
\textbf{TVM}                                                       & 254.72                \\
\textbf{Reverse*}                                                   & 604.16                \\ \hline
\textbf{Ours}                                                      & 25.34                 \\ \specialrule{.15em}{.1em}{.1em}
\end{tabular}\label{tab:10}
\end{table}

\begin{figure*}[!t]
	\centering
	    \includegraphics[width=0.95\linewidth]{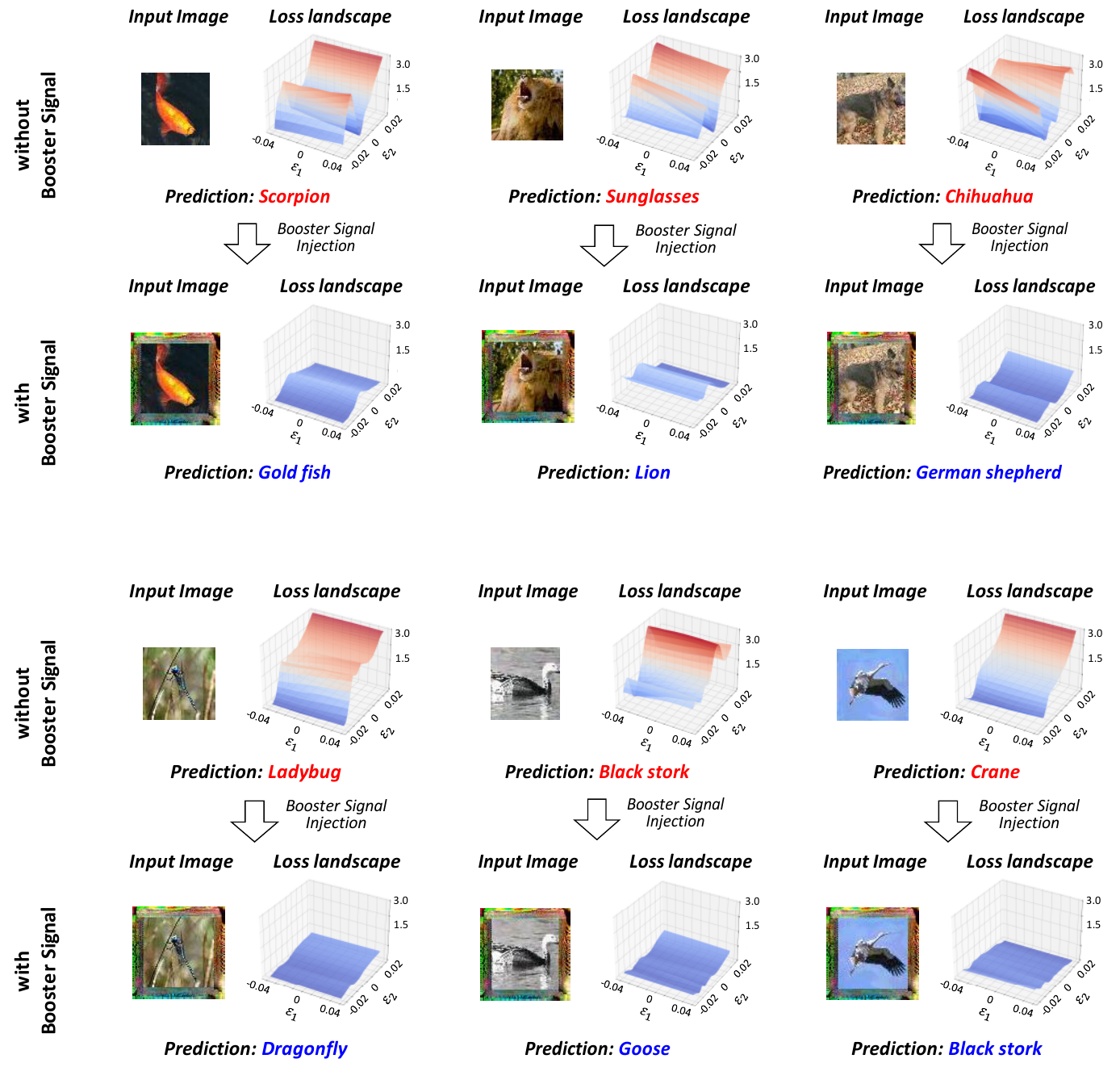}
	\caption{The loss landscape of with and without booster signal in Tiny-ImageNet. We use pretrained model trained by Madry \cite{madry2017towards}.} 

	\label{fig:5}
\end{figure*}


\subsection{Comparison with Existing Defense Methods}
\subsubsection{Defensive Performance Comparison}
There are some model-parameter agnostic adversarial defense strategies (JPEG \cite{das2017keeping}, Feature Squeeze (FS) \cite{xu2017feature}, Feature Distillation (FD) \cite{liu2019feature}, Total Variation Minimization (TVM) \cite{guo2018countering} and Reverse \cite{Mao_2021_ICCV}). To compare with those methods, we train the model by Madry \cite{madry2017towards} method then apply the model-parameter agnostic defense methods. Table \ref{tab:9} shows the defense results using existing adversarial defense strategies. As shown in the table, most of the existing defense methods cannot defend against adversarial attacks even with the adversarially trained model, since their defense strategies rely on gradient obfuscation \cite{athalye2018obfuscated}. However, our method still shows better robustness than others. Since the booster signal is optimized in an adversarial way and reduces the input gradient, it is not easily fooled by attacks. Also, since most existing methods manipulate the inside of input images, it decreases the natural accuracy.

Especially, compared with Reverse \cite{Mao_2021_ICCV} recently proposed defense methods, it decreases the natural accuracy by manipulating inside of the input images. In contrast to the Reverse, since our method injects the external signal to the outside of the image, it does not hurt the original contents which could boost natural accuracy. 
\subsubsection{Runtime Comparison}
In the main paper, we compare the existing model-parameter agnostic defense methods. Most of these methods conduct pre-processing for defense. Therefore, the execution time increases. In this section, we compare the runtime with existing model-parameter agnostic defense methods. To compare the runtime, we implement the prediction with a single A6000 GPU. Table \ref{tab:10} shows the runtime comparison of existing model-parameter agnostic defense methods. As shown in the table, since our methods simply inject the booster signal into the input image, the runtime does not increase much. Furthermore, compared to recently proposed strong defense methods (Reverse), our method shows fast runtime while it shows better defense performance.

\subsection{Analysis of Booster Signal Effect}
\subsubsection{Analysis of Loss Landscape}
Fig. \ref{fig:5} visualizes the loss landscape of randomly selected test images on the Tiny-ImageNet dataset. Following \cite{athalye2018obfuscated}, flattening the loss landscape could be evidence to support that the defense does not cause a gradient obfuscation. To visualize the loss landscape, we plot the cross-entropy loss for points surrounding two images that belong to the subspace spanned by two directions. One is random direction ($\epsilon_{1}$) and the other one is adversarial ($\textrm{sign}(\nabla_xf(x))$) direction ($\epsilon_{2}$). We use pretrained model trained by Madry \cite{madry2017towards}. As shown in the figure, injecting the booster signal to the input image flattens the loss surface, indicating the substantial defensive effect of the booster signal.

\subsubsection{Analysis of Input Gradient}
\begin{figure}[!t]
	\centering
	    \includegraphics[width=0.98\linewidth]{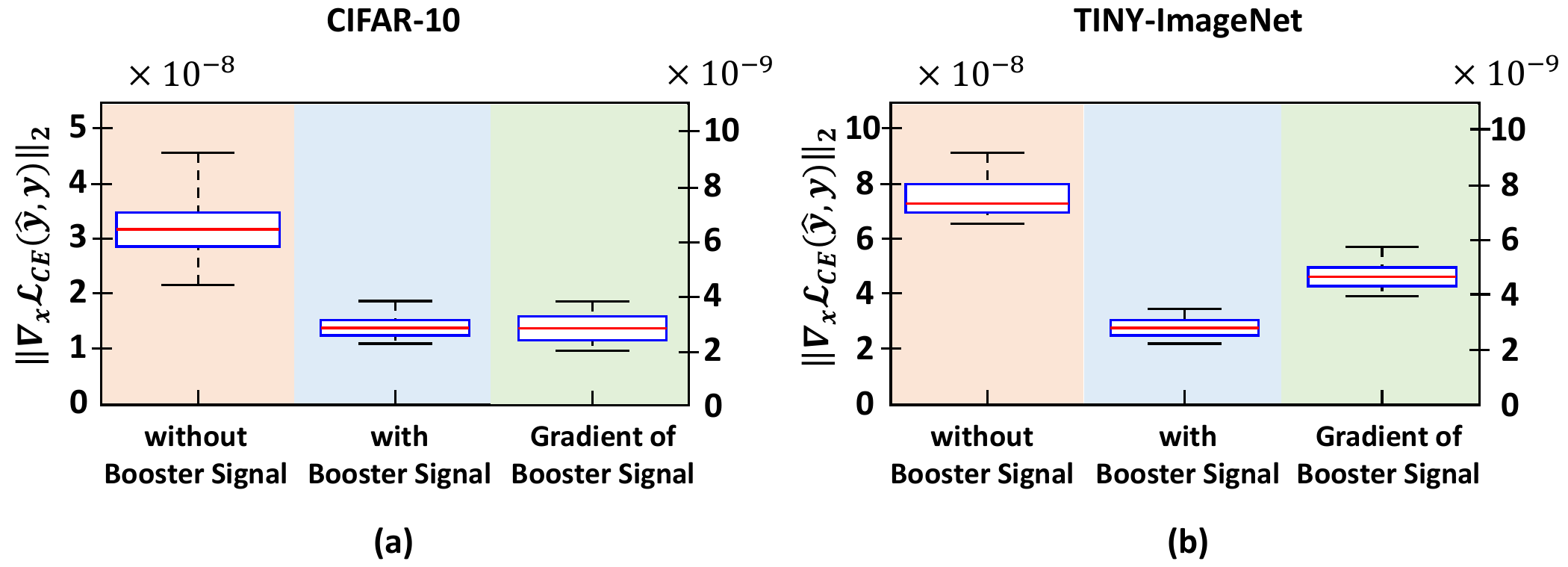}
	\caption{Distribution of L2-norm magnitudes of input gradients when a booster signal is optimized universally and apply them to all images. (a) is the input gradient distribution of CIFAR-10 and (b) is the input gradient distribution of Tiny-ImageNet. The left y-axis of (a) and (b) denotes the magnitude of input gradient for without/with booster signal. The right y-axis of (a) and (b) denotes the magnitude of input gradient for the booster signal.} 
\label{fig:6}
\end{figure}

As we discussed in Section III. B, the norm of the input gradient is related to adversarial vulnerability. Since the adversarial examples are crafted by using input gradients, smoothing the input gradients help adversarial robustness. To verify the effect of booster signal in aspect to input gradients, we statistically analyzed the input gradients of all images. Different from Fig. \ref{fig:3} that optimizes booster signals for individual images, in this section, we use a universal booster signal optimized by our proposed method. In other words, the booster signal is optimized universally for all images and apply them to all images. Fig. \ref{fig:6} shows the distribution of L2-norm magnitudes of input gradients in CIFAR-10 and Tiny-ImageNet datasets. As shown in the figure, when injecting the booster signal (`with booster signal'), it is reduced the L2-norm magnitudes of input gradients compared to 'without booster signal'. Therefore, injecting the booster signal makes the input be robust to adversarial attacks. 

Considering the analysis of the input gradient and loss landscape, it can be interpreted that injecting the booster signal can make the input itself to be robust by reducing the input gradient. Therefore, even though the booster signal is attacked, we can defend against the attack effectively.

\subsection{Effect of Number of Image Set}
\begin{figure}[!t]
	\centering
	    \includegraphics[width=0.98\linewidth]{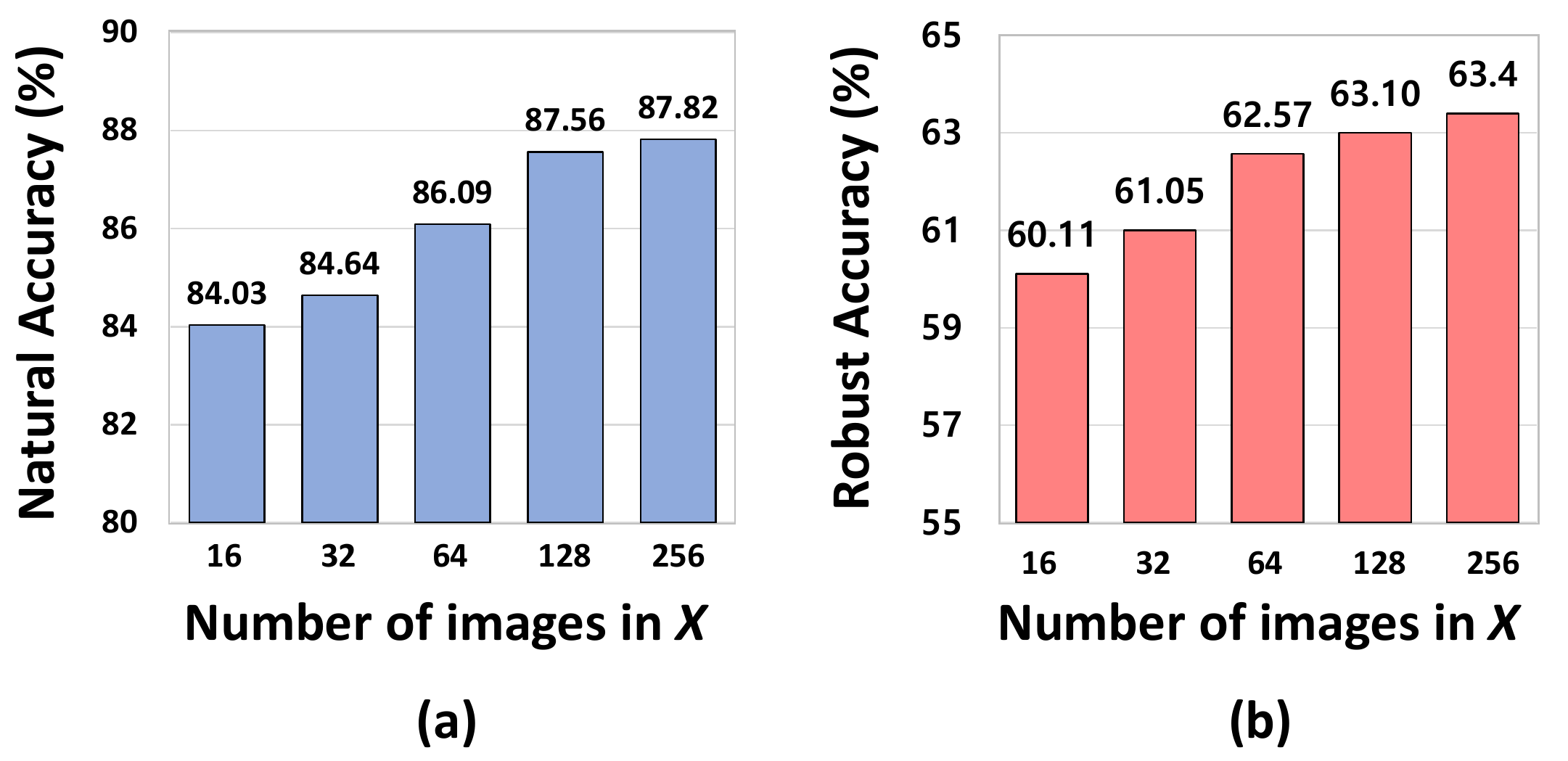}
	\caption{Natural accuracy and adversarial robustness vs. the number of images in $X$ on CIFAR-10 dataset. (a) is the natural accuracy and (b) is the robust accuracy.} 

	\label{fig:7}
\end{figure}

Fig. \ref{fig:7} shows the natural accuracy and adversarial robustness versus the number of images in $X$ on the CIFAR-10 dataset. To generate the image-agnostic booster signal, we optimize the booster signal with a subset of images sampled from the training data distribution $\mu$. Then, the booster signal makes the prediction to be correct on the data sampled from $\mu$. As shown in the figure, when the size of $X$ is small, the booster signal effect is marginal. Then, it shows similar results as standard AT results since the booster signal could not represent the data distribution. However, as the size of $X$ increases, it increases the natural accuracy and robust accuracy. It can be interpreted that, for the image-agnostic signal, a sufficient number of images in $X$ must be ensured.

\begin{table}[!t]
\centering
\caption{Comparison of the proposed method (Ours) and the result of training the original AT method (Base) for the same amount of training time as the proposed method. The experiment was conducted on WideResNet-28-10 with the CIFAR-10 dataset.}
\begin{tabular}{cccccc}
\specialrule{.15em}{.1em}{.1em}
\multicolumn{2}{c}{\textbf{Method}}    & \textbf{FGSM}  & \textbf{PGD-20} & \textbf{CW}    & \textbf{AutoAttack} \\ \hline
\multirow{2}{*}{\textbf{Madry}} & Base & 59.81          & 54.07           & 49.38          & 48.21               \\
                                & Ours & \textbf{62.45} & \textbf{57.50}  & \textbf{51.28} & \textbf{52.32}      \\ \specialrule{.15em}{.1em}{.1em}
\end{tabular}
\label{tab:11}
\end{table}

\begin{table}[!t]
\centering
\caption{Experiment results when the booster signal is randomly selected. The experiment was conducted on WideResNet28-10 with the CIFAR-10 dataset. RBS denotes the Random Booster Signal.}

\resizebox{0.98\linewidth}{!}{
\begin{tabular}{clcccc}
\specialrule{.15em}{.1em}{.1em}
\multicolumn{2}{c}{\textbf{Method}}     & \textbf{FGSM}  & \textbf{PGD-20} & \textbf{CW}    & \textbf{AutoAttack} \\ \hline
\multirow{2}{*}{\textbf{Madry}}  & Base & 59.38          & 54.21           & 49.19          & 47.93               \\
                                 & Ours+RBS & \textbf{64.21} & \textbf{59.11}  & \textbf{53.13} & \textbf{54.90}      \\ \hline
\multirow{2}{*}{\textbf{TRADES}} & Base & 61.47          & 56.24           & 50.79          & 49.85               \\
                                 & Ours+RBS & \textbf{66.85} & \textbf{60.75}  & \textbf{54.12} & \textbf{54.17}      \\ \hline
\multirow{2}{*}{\textbf{MART}}   & Base & 61.54          & 56.23           & 52.41          & 51.01               \\
                                 & Ours+RBS & \textbf{66.60} & \textbf{60.10}  & \textbf{55.83} & \textbf{55.37}      \\ \hline
\multirow{2}{*}{\textbf{GAIRAT}} & Base & 62.51          & 57.82           & 44.38          & 40.51               \\
                                 & Ours+RBS & \textbf{70.23} & \textbf{61.08}  & \textbf{46.98} & \textbf{43.56}      \\ \hline
\multirow{2}{*}{\textbf{FAT}}    & Base & 61.51          & 48.28           & 47.27          & 46.72               \\
                                 & Ours+RBS & \textbf{68.35} & \textbf{51.27}  & \textbf{50.48} & \textbf{49.31}      \\ \hline
\multirow{2}{*}{\textbf{HAT}}    & Base & 63.08          & 56.75           & 53.92          & 52.52               \\
                                 & Ours+RBS & \textbf{65.21} & \textbf{60.16}  & \textbf{56.73} & \textbf{55.98}      \\
\specialrule{.15em}{.1em}{.1em}
\end{tabular}%
}\label{tab:12}
\end{table}
\section{Discussion}
In the case of the proposed method, since we optimize not only model parameters but also optimize the booster signal, it requires additional optimization steps for the booster signal. This point can be regarded as an additional cost of the proposed method. For example, in terms of training time, the proposed method requires extra training time. In this context, we conduct experiments with the original adversarial training method at the same time as the proposed method on the CIFAR-10 dataset with WideResNet-28-10. The results are shown in Table \ref{tab:11}. Table \ref{tab:11} shows the result of training the original AT method for as much time as the proposed method is trained. As shown in the table, the performance of the original method does not change significantly even if additional training is performed, and the proposed method outperforms than original AT method.

For future directions, it will be possible to randomize the booster signal to improve robustness. If the position of the signal, the signal size, the signal value, etc are randomized so that the attacker cannot know the information about the signal, we can further improve the robustness. To verify this, we briefly conduct the experiment with a random booster signal (RBS). To this end, we generated 10 booster signals and randomly selected them at inference. The results are shown in Table \ref{tab:12}. As shown in the table, if we use a random booster signal, we could further improve the adversarial robustness.

\section{Conclusion}
In this paper, we introduce a new defense methodology with an external signal called Booster Signal. Different from previous existing adversarial training methods that handle the model parameter, our proposed method exploits the external signal other than the model parameter to improve the robustness. By injecting the booster signal into the outside of the image, it reduces the input gradient that makes the input to be robust. Also, the optimized booster signal is image agnostic. Therefore once the signal is optimized, we could inject the signal into any images. Furthermore, since the booster signal is separated from the model parameters, we can optimize the booster signal in parallel to any existing AT methods. Extensive experimental results suggest that the proposed method can improve the robustness of existing AT methods under stronger attacks and be general and flexible enough to be adopted on any AT methods.

\bibliographystyle{IEEEtran}
\bibliography{output.bbl}

\begin{thebibliography}{10}
\providecommand{\url}[1]{#1}
\csname url@samestyle\endcsname
\providecommand{\newblock}{\relax}
\providecommand{\bibinfo}[2]{#2}
\providecommand{\BIBentrySTDinterwordspacing}{\spaceskip=0pt\relax}
\providecommand{\BIBentryALTinterwordstretchfactor}{4}
\providecommand{\BIBentryALTinterwordspacing}{\spaceskip=\fontdimen2\font plus
\BIBentryALTinterwordstretchfactor\fontdimen3\font minus
  \fontdimen4\font\relax}
\providecommand{\BIBforeignlanguage}[2]{{%
\expandafter\ifx\csname l@#1\endcsname\relax
\typeout{** WARNING: IEEEtran.bst: No hyphenation pattern has been}%
\typeout{** loaded for the language `#1'. Using the pattern for}%
\typeout{** the default language instead.}%
\else
\language=\csname l@#1\endcsname
\fi
#2}}
\providecommand{\BIBdecl}{\relax}
\BIBdecl

\bibitem{he2016deep}
K.~He, X.~Zhang, S.~Ren, and J.~Sun, ``Deep residual learning for image
  recognition,'' in \emph{Proceedings of the IEEE conference on computer vision
  and pattern recognition}, 2016, pp. 770--778.

\bibitem{lee2020structure}
H.~J. Lee, J.~U. Kim, S.~Lee, H.~G. Kim, and Y.~M. Ro, ``Structure boundary
  preserving segmentation for medical image with ambiguous boundary,'' in
  \emph{Proceedings of the IEEE/CVF Conference on Computer Vision and Pattern
  Recognition}, 2020, pp. 4817--4826.

\bibitem{lin2017feature}
T.-Y. Lin, P.~Doll{\'a}r, R.~Girshick, K.~He, B.~Hariharan, and S.~Belongie,
  ``Feature pyramid networks for object detection,'' in \emph{Proceedings of
  the IEEE conference on computer vision and pattern recognition}, 2017, pp.
  2117--2125.

\bibitem{kim2019bbc}
J.~U. Kim, J.~Kwon, H.~G. Kim, and Y.~M. Ro, ``Bbc net: Bounding-box critic
  network for occlusion-robust object detection,'' \emph{IEEE transactions on
  circuits and systems for video technology}, vol.~30, no.~4, pp. 1037--1050,
  2019.

\bibitem{moritz2020streaming}
N.~Moritz, T.~Hori, and J.~Le, ``Streaming automatic speech recognition with
  the transformer model,'' in \emph{ICASSP 2020-2020 IEEE International
  Conference on Acoustics, Speech and Signal Processing (ICASSP)}.\hskip 1em
  plus 0.5em minus 0.4em\relax IEEE, 2020, pp. 6074--6078.

\bibitem{kim2021cromm}
M.~Kim, J.~Hong, S.~J. Park, and Y.~M. Ro, ``Cromm-vsr: Cross-modal memory
  augmented visual speech recognition,'' \emph{IEEE Transactions on
  Multimedia}, 2021.

\bibitem{kim2021lip}
M.~Kim, J.~Hong, and Y.~M. Ro, ``Lip to speech synthesis with visual context
  attentional gan,'' in \emph{Thirty-Fifth Conference on Neural Information
  Processing Systems}, 2021.

\bibitem{dong2018speech}
L.~Dong, S.~Xu, and B.~Xu, ``Speech-transformer: a no-recurrence
  sequence-to-sequence model for speech recognition,'' in \emph{2018 IEEE
  International Conference on Acoustics, Speech and Signal Processing
  (ICASSP)}.\hskip 1em plus 0.5em minus 0.4em\relax IEEE, 2018, pp. 5884--5888.

\bibitem{zhang2020text}
H.~Zhang and J.~Zhang, ``Text graph transformer for document classification,''
  in \emph{Conference on Empirical Methods in Natural Language Processing
  (EMNLP)}, 2020.

\bibitem{ma2019tensorized}
X.~Ma, P.~Zhang, S.~Zhang, N.~Duan, Y.~Hou, M.~Zhou, and D.~Song, ``A
  tensorized transformer for language modeling,'' \emph{Advances in Neural
  Information Processing Systems}, vol.~32, pp. 2232--2242, 2019.

\bibitem{tenney2019bert}
I.~Tenney, D.~Das, and E.~Pavlick, ``Bert rediscovers the classical nlp
  pipeline,'' \emph{arXiv preprint arXiv:1905.05950}, 2019.

\bibitem{8611298}
X.~Yuan, P.~He, Q.~Zhu, and X.~Li, ``Adversarial examples: Attacks and defenses
  for deep learning,'' \emph{IEEE Transactions on Neural Networks and Learning
  Systems}, vol.~30, no.~9, pp. 2805--2824, 2019.

\bibitem{akhtar2021advances}
N.~Akhtar, A.~Mian, N.~Kardan, and M.~Shah, ``Advances in adversarial attacks
  and defenses in computer vision: A survey,'' \emph{IEEE Access}, vol.~9, pp.
  155\,161--155\,196, 2021.

\bibitem{zhang2021adversarial}
Z.~Zhang, Z.~Zhang, Y.~Zhou, L.~Wu, S.~Wu, X.~Han, D.~Dou, T.~Che, and D.~Yan,
  ``Adversarial attack against cross-lingual knowledge graph alignment,'' in
  \emph{Proceedings of the 2021 Conference on Empirical Methods in Natural
  Language Processing}, 2021, pp. 5320--5337.

\bibitem{9302639}
J.~Liu, N.~Akhtar, and A.~Mian, ``Adversarial attack on skeleton-based human
  action recognition,'' \emph{IEEE Transactions on Neural Networks and Learning
  Systems}, pp. 1--14, 2020.

\bibitem{goodfellow2014explaining}
I.~J. Goodfellow, J.~Shlens, and C.~Szegedy, ``Explaining and harnessing
  adversarial examples,'' \emph{arXiv preprint arXiv:1412.6572}, 2014.

\bibitem{madry2017towards}
A.~Madry, A.~Makelov, L.~Schmidt, D.~Tsipras, and A.~Vladu, ``Towards deep
  learning models resistant to adversarial attacks,'' \emph{arXiv preprint
  arXiv:1706.06083}, 2017.

\bibitem{carlini2017towards}
N.~Carlini and D.~Wagner, ``Towards evaluating the robustness of neural
  networks,'' in \emph{2017 ieee symposium on security and privacy (sp)}.\hskip
  1em plus 0.5em minus 0.4em\relax IEEE, 2017, pp. 39--57.

\bibitem{das2017keeping}
N.~Das, M.~Shanbhogue, S.-T. Chen, F.~Hohman, L.~Chen, M.~E. Kounavis, and
  D.~H. Chau, ``Keeping the bad guys out: Protecting and vaccinating deep
  learning with jpeg compression,'' \emph{arXiv preprint arXiv:1705.02900},
  2017.

\bibitem{liu2019feature}
Z.~Liu, Q.~Liu, T.~Liu, N.~Xu, X.~Lin, Y.~Wang, and W.~Wen, ``Feature
  distillation: Dnn-oriented jpeg compression against adversarial examples,''
  in \emph{2019 IEEE/CVF Conference on Computer Vision and Pattern Recognition
  (CVPR)}.\hskip 1em plus 0.5em minus 0.4em\relax IEEE, 2019, pp. 860--868.

\bibitem{naseer2020self}
M.~Naseer, S.~Khan, M.~Hayat, F.~S. Khan, and F.~Porikli, ``A self-supervised
  approach for adversarial robustness,'' in \emph{Proceedings of the IEEE/CVF
  Conference on Computer Vision and Pattern Recognition}, 2020, pp. 262--271.

\bibitem{Mao_2021_ICCV}
C.~Mao, M.~Chiquier, H.~Wang, J.~Yang, and C.~Vondrick, ``Adversarial attacks
  are reversible with natural supervision,'' in \emph{Proceedings of the
  IEEE/CVF International Conference on Computer Vision (ICCV)}, October 2021,
  pp. 661--671.

\bibitem{naseer2019local}
M.~Naseer, S.~Khan, and F.~Porikli, ``Local gradients smoothing: Defense
  against localized adversarial attacks,'' in \emph{2019 IEEE Winter Conference
  on Applications of Computer Vision (WACV)}.\hskip 1em plus 0.5em minus
  0.4em\relax IEEE, 2019, pp. 1300--1307.

\bibitem{raff2019barrage}
E.~Raff, J.~Sylvester, S.~Forsyth, and M.~McLean, ``Barrage of random
  transforms for adversarially robust defense,'' in \emph{Proceedings of the
  IEEE/CVF Conference on Computer Vision and Pattern Recognition}, 2019, pp.
  6528--6537.

\bibitem{xie2017mitigating}
C.~Xie, J.~Wang, Z.~Zhang, Z.~Ren, and A.~Yuille, ``Mitigating adversarial
  effects through randomization,'' \emph{arXiv preprint arXiv:1711.01991},
  2017.

\bibitem{lee2020robust}
H.~Lee, H.~J. Lee, S.~T. Kim, and Y.~M. Ro, ``Robust ensemble model training
  via random layer sampling against adversarial attack,'' \emph{arXiv preprint
  arXiv:2005.10757}, 2020.

\bibitem{liu2018towards}
X.~Liu, M.~Cheng, H.~Zhang, and C.-J. Hsieh, ``Towards robust neural networks
  via random self-ensemble,'' in \emph{Proceedings of the European Conference
  on Computer Vision (ECCV)}, 2018, pp. 369--385.

\bibitem{athalye2018obfuscated}
A.~Athalye, N.~Carlini, and D.~Wagner, ``Obfuscated gradients give a false
  sense of security: Circumventing defenses to adversarial examples,'' in
  \emph{International conference on machine learning}.\hskip 1em plus 0.5em
  minus 0.4em\relax PMLR, 2018, pp. 274--283.

\bibitem{Akhtar_2018_CVPR}
N.~Akhtar, J.~Liu, and A.~Mian, ``Defense against universal adversarial
  perturbations,'' in \emph{Proceedings of the IEEE Conference on Computer
  Vision and Pattern Recognition (CVPR)}, June 2018.

\bibitem{pmlr-v139-zhao21e}
\BIBentryALTinterwordspacing
X.~Zhao, Z.~Zhang, Z.~Zhang, L.~Wu, J.~Jin, Y.~Zhou, R.~Jin, D.~Dou, and
  D.~Yan, ``Expressive 1-lipschitz neural networks for robust multiple graph
  learning against adversarial attacks,'' in \emph{Proceedings of the 38th
  International Conference on Machine Learning}, ser. Proceedings of Machine
  Learning Research, M.~Meila and T.~Zhang, Eds., vol. 139.\hskip 1em plus
  0.5em minus 0.4em\relax PMLR, 18--24 Jul 2021, pp. 12\,719--12\,735.
  [Online]. Available: \url{https://proceedings.mlr.press/v139/zhao21e.html}
\BIBentrySTDinterwordspacing

\bibitem{srinivasan2021robustifying}
V.~Srinivasan, C.~Rohrer, A.~Marban, K.-R. M{\"u}ller, W.~Samek, and
  S.~Nakajima, ``Robustifying models against adversarial attacks by langevin
  dynamics,'' \emph{Neural Networks}, vol. 137, pp. 1--17, 2021.

\bibitem{9466420}
Q.~Liu and W.~Wen, ``Model compression hardens deep neural networks: A new
  perspective to prevent adversarial attacks,'' \emph{IEEE Transactions on
  Neural Networks and Learning Systems}, pp. 1--12, 2021.

\bibitem{bai2021recent}
T.~Bai, J.~Luo, J.~Zhao, B.~Wen, and Q.~Wang, ``Recent advances in adversarial
  training for adversarial robustness,'' \emph{arXiv preprint
  arXiv:2102.01356}, 2021.

\bibitem{zhang2021geometryaware}
\BIBentryALTinterwordspacing
J.~Zhang, J.~Zhu, G.~Niu, B.~Han, M.~Sugiyama, and M.~Kankanhalli,
  ``Geometry-aware instance-reweighted adversarial training,'' in
  \emph{International Conference on Learning Representations}, 2021. [Online].
  Available: \url{https://openreview.net/forum?id=iAX0l6Cz8ub}
\BIBentrySTDinterwordspacing

\bibitem{zhang2020attacks}
J.~Zhang, X.~Xu, B.~Han, G.~Niu, L.~Cui, M.~Sugiyama, and M.~Kankanhalli,
  ``Attacks which do not kill training make adversarial learning stronger,'' in
  \emph{International Conference on Machine Learning}.\hskip 1em plus 0.5em
  minus 0.4em\relax PMLR, 2020, pp. 11\,278--11\,287.

\bibitem{Wang2020Improving}
\BIBentryALTinterwordspacing
Y.~Wang, D.~Zou, J.~Yi, J.~Bailey, X.~Ma, and Q.~Gu, ``Improving adversarial
  robustness requires revisiting misclassified examples,'' in
  \emph{International Conference on Learning Representations}, 2020. [Online].
  Available: \url{https://openreview.net/forum?id=rklOg6EFwS}
\BIBentrySTDinterwordspacing

\bibitem{pmlr-v97-zhang19p}
\BIBentryALTinterwordspacing
H.~Zhang, Y.~Yu, J.~Jiao, E.~Xing, L.~E. Ghaoui, and M.~Jordan, ``Theoretically
  principled trade-off between robustness and accuracy,'' in \emph{Proceedings
  of the 36th International Conference on Machine Learning}, ser. Proceedings
  of Machine Learning Research, K.~Chaudhuri and R.~Salakhutdinov, Eds.,
  vol.~97.\hskip 1em plus 0.5em minus 0.4em\relax PMLR, 09--15 Jun 2019, pp.
  7472--7482. [Online]. Available:
  \url{https://proceedings.mlr.press/v97/zhang19p.html}
\BIBentrySTDinterwordspacing

\bibitem{kannan2018adversarial}
H.~Kannan, A.~Kurakin, and I.~Goodfellow, ``Adversarial logit pairing,''
  \emph{arXiv preprint arXiv:1803.06373}, 2018.

\bibitem{rade2021helper}
R.~Rade and S.-M. Moosavi-Dezfooli, ``Helper-based adversarial training:
  Reducing excessive margin to achieve a better accuracy vs. robustness
  trade-off,'' in \emph{ICML 2021 Workshop on Adversarial Machine Learning},
  2021.

\bibitem{li2020vulnerability}
C.~Li, H.~Tang, C.~Deng, L.~Zhan, and W.~Liu, ``Vulnerability vs. reliability:
  Disentangled adversarial examples for cross-modal learning,'' in
  \emph{Proceedings of the 26th ACM SIGKDD International Conference on
  Knowledge Discovery \& Data Mining}, 2020, pp. 421--429.

\bibitem{9308597}
H.~Liu and G.~Ditzler, ``Adversarial audio attacks that evade temporal
  dependency,'' in \emph{2020 IEEE Symposium Series on Computational
  Intelligence (SSCI)}, 2020, pp. 639--646.

\bibitem{srinivasan2019black}
V.~Srinivasan, E.~E. Kuruoglu, K.-R. M{\"u}ller, W.~Samek, and S.~Nakajima,
  ``Black-box decision based adversarial attack with symmetric $\alpha$-stable
  distribution,'' in \emph{2019 27th European Signal Processing Conference
  (EUSIPCO)}.\hskip 1em plus 0.5em minus 0.4em\relax IEEE, 2019, pp. 1--5.

\bibitem{duan2021advdrop}
R.~Duan, Y.~Chen, D.~Niu, Y.~Yang, A.~K. Qin, and Y.~He, ``Advdrop: Adversarial
  attack to dnns by dropping information,'' in \emph{Proceedings of the
  IEEE/CVF International Conference on Computer Vision}, 2021, pp. 7506--7515.

\bibitem{wang2020hamiltonian}
H.~Wang, G.~Li, X.~Liu, and L.~Lin, ``A hamiltonian monte carlo method for
  probabilistic adversarial attack and learning,'' \emph{IEEE Transactions on
  Pattern Analysis and Machine Intelligence}, 2020.

\bibitem{croce2020reliable}
F.~Croce and M.~Hein, ``Reliable evaluation of adversarial robustness with an
  ensemble of diverse parameter-free attacks,'' in \emph{International
  conference on machine learning}.\hskip 1em plus 0.5em minus 0.4em\relax PMLR,
  2020, pp. 2206--2216.

\bibitem{croce2020minimally}
------, ``Minimally distorted adversarial examples with a fast adaptive
  boundary attack,'' in \emph{International Conference on Machine
  Learning}.\hskip 1em plus 0.5em minus 0.4em\relax PMLR, 2020, pp. 2196--2205.

\bibitem{andriushchenko2020square}
M.~Andriushchenko, F.~Croce, N.~Flammarion, and M.~Hein, ``Square attack: a
  query-efficient black-box adversarial attack via random search,'' in
  \emph{European Conference on Computer Vision}.\hskip 1em plus 0.5em minus
  0.4em\relax Springer, 2020, pp. 484--501.

\bibitem{croce2021robustbench}
\BIBentryALTinterwordspacing
F.~Croce, M.~Andriushchenko, V.~Sehwag, E.~Debenedetti, N.~Flammarion,
  M.~Chiang, P.~Mittal, and M.~Hein, ``Robustbench: a standardized adversarial
  robustness benchmark,'' in \emph{Thirty-fifth Conference on Neural
  Information Processing Systems Datasets and Benchmarks Track (Round 2)},
  2021. [Online]. Available: \url{https://openreview.net/forum?id=SSKZPJCt7B}
\BIBentrySTDinterwordspacing

\bibitem{aprilpyone2021block}
M.~AprilPyone and H.~Kiya, ``Block-wise image transformation with secret key
  for adversarially robust defense,'' \emph{IEEE Transactions on Information
  Forensics and Security}, vol.~16, pp. 2709--2723, 2021.

\bibitem{meng2017magnet}
D.~Meng and H.~Chen, ``Magnet: a two-pronged defense against adversarial
  examples,'' in \emph{Proceedings of the 2017 ACM SIGSAC conference on
  computer and communications security}, 2017, pp. 135--147.

\bibitem{song2018pixeldefend}
\BIBentryALTinterwordspacing
Y.~Song, T.~Kim, S.~Nowozin, S.~Ermon, and N.~Kushman, ``Pixeldefend:
  Leveraging generative models to understand and defend against adversarial
  examples,'' in \emph{International Conference on Learning Representations},
  2018. [Online]. Available: \url{https://openreview.net/forum?id=rJUYGxbCW}
\BIBentrySTDinterwordspacing

\bibitem{ross2018improving}
A.~S. Ross and F.~Doshi-Velez, ``Improving the adversarial robustness and
  interpretability of deep neural networks by regularizing their input
  gradients,'' in \emph{Thirty-second AAAI conference on artificial
  intelligence}, 2018.

\bibitem{chan2020thinks}
A.~Chan, Y.~Tay, and Y.-S. Ong, ``What it thinks is important is important:
  Robustness transfers through input gradients,'' in \emph{Proceedings of the
  IEEE/CVF Conference on Computer Vision and Pattern Recognition}, 2020, pp.
  332--341.

\bibitem{krizhevsky2009learning}
A.~Krizhevsky, G.~Hinton \emph{et~al.}, ``Learning multiple layers of features
  from tiny images,'' 2009.

\bibitem{le2015tiny}
Y.~Le and X.~Yang, ``Tiny imagenet visual recognition challenge,'' \emph{CS
  231N}, vol.~7, no.~7, p.~3, 2015.

\bibitem{russakovsky2015imagenet}
O.~Russakovsky, J.~Deng, H.~Su, J.~Krause, S.~Satheesh, S.~Ma, Z.~Huang,
  A.~Karpathy, A.~Khosla, M.~Bernstein \emph{et~al.}, ``Imagenet large scale
  visual recognition challenge,'' \emph{International journal of computer
  vision}, vol. 115, no.~3, pp. 211--252, 2015.

\bibitem{Wong2020Fast}
\BIBentryALTinterwordspacing
E.~Wong, L.~Rice, and J.~Z. Kolter, ``Fast is better than free: Revisiting
  adversarial training,'' in \emph{International Conference on Learning
  Representations}, 2020. [Online]. Available:
  \url{https://openreview.net/forum?id=BJx040EFvH}
\BIBentrySTDinterwordspacing

\bibitem{zagoruyko2016wide}
S.~Zagoruyko and N.~Komodakis, ``Wide residual networks,'' \emph{arXiv preprint
  arXiv:1605.07146}, 2016.

\bibitem{xu2017feature}
W.~Xu, D.~Evans, and Y.~Qi, ``Feature squeezing: Detecting adversarial examples
  in deep neural networks,'' \emph{arXiv preprint arXiv:1704.01155}, 2017.

\bibitem{guo2018countering}
\BIBentryALTinterwordspacing
C.~Guo, M.~Rana, M.~Cisse, and L.~van~der Maaten, ``Countering adversarial
  images using input transformations,'' in \emph{International Conference on
  Learning Representations}, 2018. [Online]. Available:
  \url{https://openreview.net/forum?id=SyJ7ClWCb}
\BIBentrySTDinterwordspacing

\end{thebibliography}

\vfill

\begin{IEEEbiography}[{\includegraphics[width=1in,height=1.25in,clip]{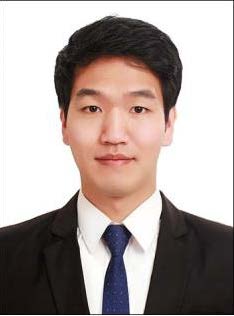}}]{HONG JOO LEE}
received the B.S. degree from Ajou University, Suwon, South Korea, in 2016, and the M.S. and Ph.D. degrees from the Korea Advanced Institute of Science and Technology (KAIST), Daejeon, South Korea, in 2018 and 2023. His research interests include deep learning, machine learning, medical image segmentation, and adversarial robustness.
\end{IEEEbiography}

\begin{IEEEbiography}[{\includegraphics[width=1in,height=1.25in,clip]{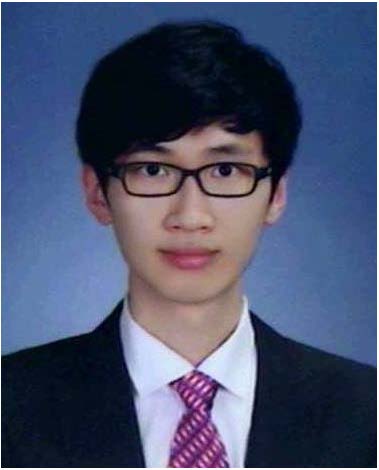}}]{YOUNGJOON YU}
received the B.S. degree in electrical engineering from Korea Advanced Institute of Science and Technology (KAIST), Daejeon, South Korea in 2013, and the M.S. degree in the management engineering from KAIST in 2017. He is currently pursuing the Ph.D. in electrical engineering at KAIST, Daejeon, South Korea. His research interests include deep learning, multi-sensor learning, and adversarial robustness.
\end{IEEEbiography}

\begin{IEEEbiography}[{\includegraphics[width=1in,height=1.25in,clip]{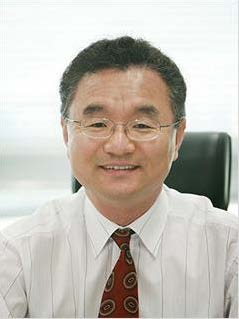}}]{YONG MAN RO}
(Senior Member, IEEE) received a B.S. degree from Yonsei University, Seoul, South Korea, and a M.S. and Ph.D. degrees from the Korea Advanced Institute of Science and Technology (KAIST), Daejeon, South Korea. He was a Researcher at Columbia University, a Visiting Researcher at the University of California at Irvine, Irvine, CA, USA, and a Research Fellow of the University of California at Berkeley, Berkeley, CA, USA. He was a Visiting Professor with the Department of Electrical and Computer Engineering, University of Toronto, Canada. He is currently a Professor at the Department of Electrical Engineering and the Director of the Center for Applied Research in Artificial Intelligence (CARAI), KAIST. Among the years, he has been conducting research in a wide spectrum of image and video systems research topics. Among those topics, his interests include image processing, computer vision, visual recognition, multimodal learning, video representation/compression, and object detection. He received the Young Investigator Finalist Award of ISMRM, in 1992, and the Year’s Scientist Award (Korea), in 2003. He served as an Associate Editor for IEEE SIGNAL PROCESSING LETTERS. He currently serves as an Associate Editor for IEEE TRANSACTIONS ON CIRCUITS AND SYSTEMS FOR VIDEO TECHNOLOGY. He served as a TPC in many international conferences, including the program chair, and organized special sessions.
\end{IEEEbiography}

\end{document}